\documentclass{article} %
\usepackage{iclr2025_conference,times}

\usepackage{xcolor}
\definecolor{citecolor}{HTML}{2779af}
\definecolor{linkcolor}{HTML}{c0392b}
\usepackage[pagebackref=false,breaklinks=true,colorlinks,bookmarks=false,citecolor=citecolor,linkcolor=linkcolor]{hyperref}

\usepackage{url}
\usepackage{booktabs}
\usepackage{multirow}
\usepackage{xspace}
\usepackage{graphicx}
\usepackage{tablefootnote}
\usepackage{amssymb}
\usepackage{amsmath}
\usepackage{wrapfig}
\usepackage{nicefrac}
\usepackage{inconsolata}
\usepackage[capitalise]{cleveref}
\usepackage[group-separator={,},group-minimum-digits=3,reset-text-series=false,text-series-to-math=true,unit-mode=text]{siunitx}
\usepackage{enumitem}
\usepackage{verbatim}

\usepackage[format=hang,singlelinecheck=false]{subcaption}

\usepackage{tabularx}
\usepackage{algorithm}
\usepackage[indLines=true,noEnd=false,italicComments=false,commentColor=black]{algpseudocodex}

\usepackage{array,etoolbox}
\preto\tabular{\setcounter{magicrownumbers}{0}}
\newcounter{magicrownumbers}
\def\rownumber{}

\crefname{equation}{Equation}{Equations}

\def\reffig#1{\cref{fig:#1}}
\def\refalg#1{\cref{alg:#1}}

\def\refsec#1{\cref{sec:#1}}

\def\lblfig#1{\label{fig:#1}}
\def\lblalg#1{\label{alg:#1}}
\def\lbleq#1{\label{eq:#1}}
\def\lblsec#1{\label{sec:#1}}

\newcommand{\mat}[1]{\mathbf{#1}}
\renewcommand{\vec}[1]{\mathbf{#1}}
\newcommand{\imat}[1]{#1}
\newcommand{\ivec}[1]{#1}
\newcommand{\xhdr}[1]{\vspace{0pt}\noindent\textbf{#1}\xspace}
\newcommand{\softmax}{\mathrm{softmax}}
\newcommand{\pluseq}{\mathrel{+}=}

\definecolor{ecolor}{HTML}{f39c12}
\newcommand{\mE}{{\color{ecolor}E}}

\definecolor{ccolor}{HTML}{2980b9}
\newcommand{\mC}{{\color{ccolor}C}}

\definecolor{lsecolor}{HTML}{c0392b}
\newcommand{\mLSE}{\mathrm{\color{lsecolor}LSE}}

\definecolor{celcolor}{HTML}{e74c3c}
\newcommand{\mCEL}{\mathrm{\color{celcolor}CEL}}

\definecolor{acolor}{HTML}{d35400}

\def\V{{|V|}}
\def\BN{{N_B}}
\def\BV{{V_B}}
\def\BD{{D_B}}
\def\dim#1#2{{{#1} \times {#2}}}

\NewDocumentCommand{\todo}{om}{\textcolor{red}{Todo\IfValueT{#1}{ (#1)}: #2}}

\title{Cut Your Losses \\in Large-Vocabulary Language Models}

\author{
Erik Wijmans\thanks{Corresponding author: \texttt{ewijmans@apple.com}}~~
Brody Huval~~
Alexander Hertzberg~~
Vladlen Koltun~~
Philipp Kr\"ahenb\"uhl \\
\multicolumn{1}{c}{Apple}
}

\newcommand{\eg}{e.g.,\xspace}
\newcommand{\ie}{i.e.,\xspace}
\newcommand{\wrt}{wrt.\xspace}
\newcommand{\etc}{\textit{etc.}\xspace}
\newcommand{\ours}{CCE\xspace}
\newcommand{\oursfullc}{CCE-Kahan-FullC\xspace}
\newcommand{\oursfulle}{CCE-Kahan-FullE\xspace}
\newcommand{\ourskahan}{CCE-Kahan\xspace}

\iclrfinalcopy %

\begin{document}

\maketitle

\begin{abstract}
As language models grow ever larger, so do their vocabularies.
This has shifted the memory footprint of LLMs during training disproportionately to one single layer: the cross-entropy in the loss computation.
Cross-entropy builds up a logit matrix with entries for each pair of input tokens and vocabulary items and, for small models, consumes an order of magnitude more memory than the rest of the LLM combined.
We propose Cut Cross-Entropy (\ours), a method that computes the cross-entropy loss without materializing the logits for all tokens into global memory.
Rather, \ours only computes the logit for the correct token and evaluates the log-sum-exp over all logits on the fly.
We implement a custom kernel that performs the matrix multiplications and the log-sum-exp reduction over the vocabulary in flash memory, making global memory consumption for the cross-entropy computation negligible. This has a dramatic effect. Taking the Gemma 2 (2B) model as an example, \ours reduces the memory footprint of the loss computation from 24 GB to 1 MB, and the total training-time memory consumption of the classifier head from 28 GB to 1 GB.
To improve the throughput of \ours, we leverage the inherent sparsity of softmax and propose to skip elements of the gradient computation that have a negligible (i.e., below numerical precision) contribution to the gradient.
Experiments demonstrate that the dramatic reduction in memory consumption is accomplished without sacrificing training speed or convergence.

\ificlrfinal
\centerline{\url{https://github.com/apple/ml-cross-entropy}}
\fi
\end{abstract}

\section{Introduction}

Progress in large language models (LLMs) has been fueled in part by an increase in parameter count, context length, and vocabulary size (the number of tokens that can be used to represent the input).
As LLMs grew, so did the associated infrastructure.
Large mini-batch gradient descent~\citep{goyal2017large} combined with data-parallelism~\citep{hillis1986data} enabled the harnessing of increasing computational power.
ZeRO~\citep{zero2020} broke the dependence between the number of GPUs and the memory used for model parameters, gradients, and optimizer state.
Activation checkpointing~\citep{chen2016checkpointing} reduced the amount of memory used for activations, supporting the development of deeper models.
FlashAttention~\citep{flashattention} reduced the memory used in self-attention from $O(N^2)$ to $O(N)$, thereby supporting longer context windows.
These improvements gradually shifted the memory consumption of LLM training to one single layer~-- the cross-entropy loss, whose memory footprint grows with the product of vocabulary size and number of tokens per batch.
The cross-entropy loss is responsible for up to $90\%$ of the memory footprint of modern LLM training (see \reffig{teaser_regular}).
The problem grows only more acute with time, since even the  largest contemporary vocabularies (\eg 256K tokens) may benefit from further expansion~\citep{tao2024scalinglawsvocabularylarger}.

We propose a cross-entropy implementation, Cut Cross-Entropy (\ours), that has a negligible memory footprint and scales to arbitrarily large vocabularies.
Our key insight is that computation of the loss and its gradient only depends on a single log-probability, that of the ground-truth label.
With an arithmetic reformulation, we decompose the cross-entropy loss into an index matrix multiplication over a single ground-truth label and a log-sum-exp operation over all vocabulary entries for each token.
Each operation has small and well-defined inputs~-- the network embeddings and classifier matrix~-- and a single scalar output per token.
Both operations do, however, rely on a large intermediate logit matrix that computes the score for each token and potential vocabulary entry.
We show that there is no need to materialize this logit matrix in GPU memory.
Instead, we compute logits as needed in SRAM in a series of custom CUDA kernels.
The result is a cross-entropy computation that has negligible memory footprint, with no detrimental effect on latency or convergence.
See \reffig{teaser_ours} for a breakdown of memory savings and consequent batch size increases afforded by \ours.

\begin{figure}
    \captionsetup[subfigure]{justification=centering}
    \centering
    \begin{subfigure}[t]{.29\linewidth}
        \includegraphics[width=\linewidth,page=7]{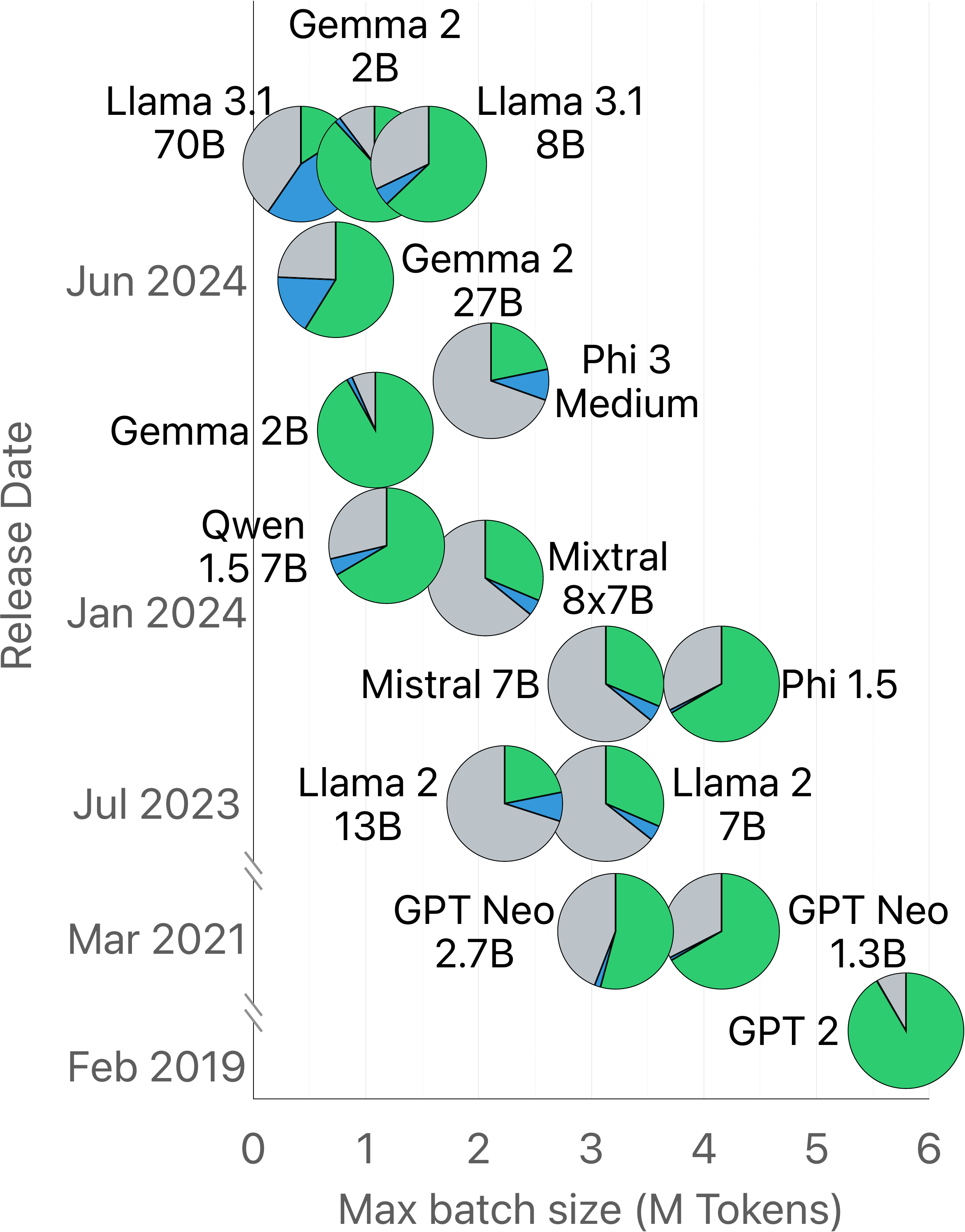}
        \caption{Regular cross-entropy}
        \lblfig{teaser_regular}
      \end{subfigure}\hfill
      \begin{subfigure}[t]{.69\linewidth}
        \includegraphics[width=\linewidth,page=8]{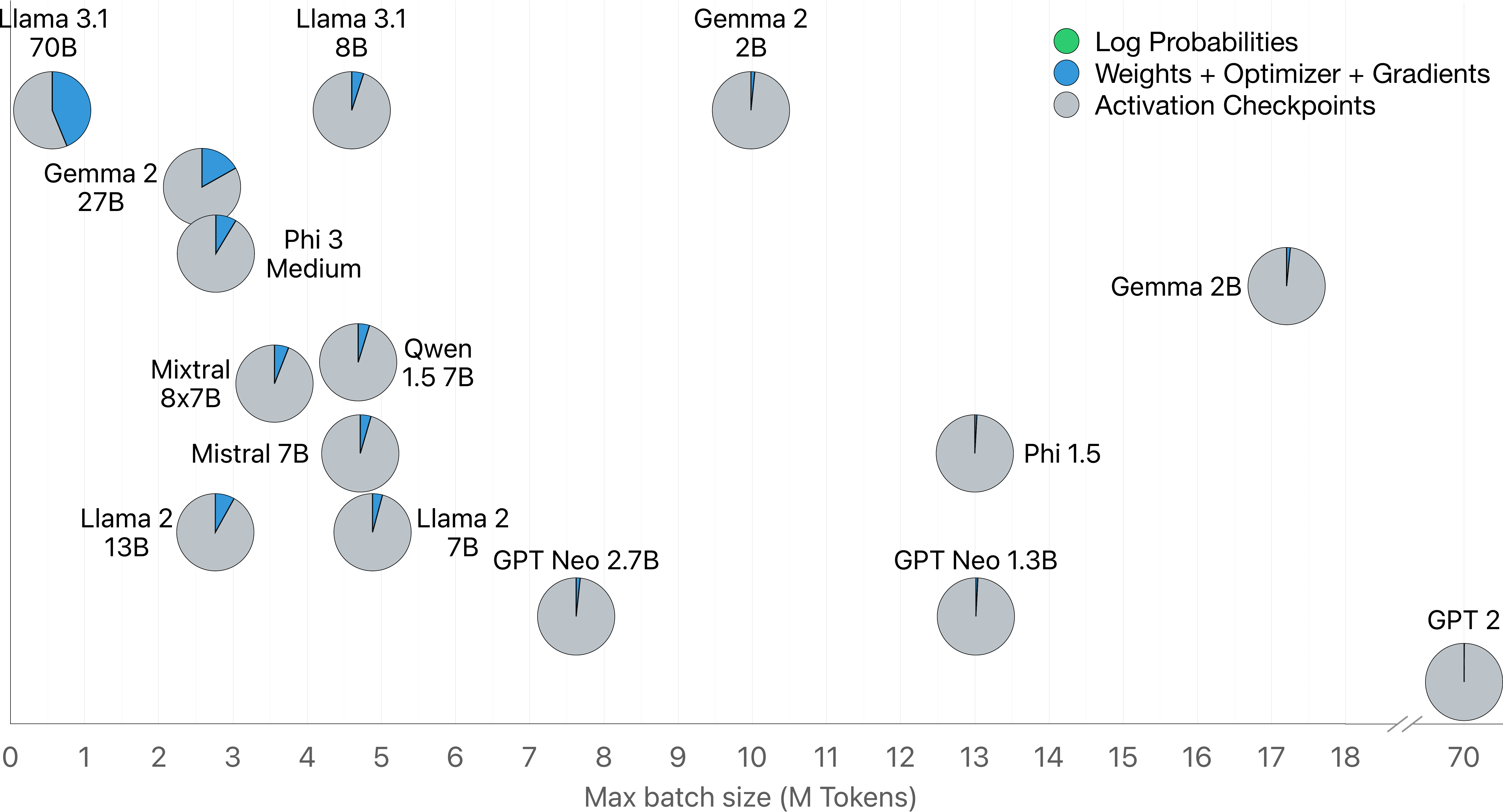}
        \caption{Cut cross-entropy (ours)}
        \lblfig{teaser_ours}
      \end{subfigure}
    \centering
    \caption{
        Memory use and maximum attainable batch size (in millions of tokens) for a variety of frontier models on a 16-GPU (80 GB each) fully-sharded data-parallel setup~\citep{zero2020} with activation checkpointing~\citep{chen2016checkpointing} and a mixed-precision 16-bit (fp16/bf16) AdamW optimizer~\citep{kingma2015adam,loshchilov2019adamw}.
        For each model, we break its memory use down into weights and optimizer states, activation checkpoints, and the log-probabilities computed by the cross-entropy loss layer.
        Our Cut Cross-Entropy (\ours) enables increasing the batch size by 1.5x (Llama 2 13B) to 10x (GPT 2, Gemma 2 2B), with no sacrifice in speed or convergence. Exact values in \cref{tab:teaser-data}.
    }
    \lblfig{teaser}
\end{figure}

\section{Related Work}

\xhdr{Attention mechanisms.}
The effectiveness of transformers~\citep{transformer2017} in modeling language has drawn attention to their compute and memory requirements.
Multiple works have proposed alternatives to scaled dot-product attention that reduce transformers' computation and memory~\citep{kitaev2020reformer,wang2020linformerselfattentionlinearcomplexity,choromanski2020rethinking}. Other model classes, such as structured state-space models~\citep{gu2022s4,gu2023mamba}, have also shown promising results. We study a different part of the model~-- its classifier head~-- that is not considered in these works.

\xhdr{Attention implementations.}
In addition to alternative attention mechanisms, the community has also tackled the daunting memory consumption of LLMs via efficient implementations.
\citet{rabe2021memeff} developed a self-attention implementation that makes use of chunking.
\citet{chen2023speed} proposed an implementation that broke the operation into two stages, reduction and matrix multiplication. This makes efficient use of GPU memory and registers but requires recomputation in the forward pass.
FlashAttention~\citep{flashattention} uses an online softmax~\citep{milakov2018softmax} and, like \ours, materializes blocks of the $N^2$-sized self-attention matrix in on-chip SRAM rather than slower global DRAM. This is one of the key ideas that \ours builds on to develop a memory-efficient cross-entropy formulation.

\xhdr{Vocabulary reduction.}
One way to minimize the amount of memory used by the log-probabilities over the tokens is to reduce the number of `active' tokens in the vocabulary. \citet{joulin2017efficient} proposed to use a vocabulary with a hierarchical structure, thereby requiring the log-probabilities for only a subset of the vocabulary at any given time. \citet{yu2023megabyte} explore tokenization-free byte-level models that operate on dramatically smaller vocabularies.

\xhdr{Sequence and model parallelism.} Sequence parallelism~\citep{deepspeed-ulysses,li2023seqparallel} enables training very large models (with large vocabularies) by splitting an individual input sequence across multiple GPUs. Various model parallelism techniques~\citep{huang2019gpipe,narayanan20139pipedream,megatronlm} achieve the same goal of training very large models (with large vocabularies) by distributing the computation and memory consumption of different pieces across multiple GPUs. 

\xhdr{Efficient cross-entropy implementations.}
A number of recent implementations use chunking to reduce the memory usage of the cross-entropy layer.
Yet chunking induces a trade-off. Memory footprint is minimized when the number of chunks is high, but latency is minimized when the number of chunks is low. \ours utilizes only on-chip SRAM and minimizes both memory footprint and latency.
Liger Kernels~\citep{liger2024} make efficient use of the GPU via chunking and by computing the loss+gradient simultaneously. The latter requires that any transform applied to the loss (such as masking) is implemented in the kernel itself. \ours has separate forward and backward stages, enabling user-defined transformations on the loss.

\section{Preliminaries}

Let $P(x) = \prod_{i=1}^N P(x_i \mid x_1 \ldots x_{i-1})$ be a Large Language Model (LLM) over a vocabulary $V$.
The LLM parameterizes an autoregressive distribution over all possible tokens $x_i \in V$ given the preceding $N-1$ tokens.
Specifically, this distribution is the combination of a backbone network ${f: x_1\ldots x_{i-1} \to \mathbb{R}^D}$ and a linear classifier $\mat{\mC} \in \mathbb{R}^\dim{D}{|V|}$:
\begin{align}
P(x_i \mid x_1 \ldots x_{i-1}) & = \softmax_{x_i}(\mat{\mC}^\top f(x_1 \ldots x_{i-1})), \\
\softmax_k(\vec v) &=  \frac{\exp(v_k)}{\sum_{j} \exp(v_j)}.
\end{align}
The backbone network $f(x_1, \ldots, x_{i-1}) \in \mathbb{R}^D$ encodes a token sequence in the $D$-dimensional feature vector.
The linear classifier $\mat{\mC} \in \mathbb{R}^\dim{D}{|V|}$ projects the embedding into an output space of the vocabulary $V$.
The $\softmax_k(\vec v)$ produces the probability over all vocabulary entries from the unnormalized log probabilities (logits) produced by $\mat{\mC}^\top f(x_1 \ldots x_{i-1})$.

\subsection{Vocabulary}

LLMs represent their input (and output) as a set of tokens in a vocabulary $V$. The vocabulary is typically constructed by a method such as Byte Pair Encoding (BPE)~\citep{gage1994bpe}. BPE initializes the vocabulary with all valid byte sequences from a standard text encoding, such as utf-8. Then, over a large corpus of text, BPE finds the most frequent pair of tokens and creates a new token that represents this pair. This continues iteratively until the maximum number of tokens is reached.

Large vocabularies enable a single token to represent multiple characters. This reduces the length of both input and output sequences, compresses larger and more diverse documents into shorter context windows, thus improving the model's comprehension while reducing computational demands.

\subsection{Inference and Training}

Even with a large vocabulary, sampling from an LLM is memory-efficient at inference time. Specifically,
the LLM produces one token at a time, computing $P(x_i | x_1 \ldots x_{i-1})$ and sampling from this distribution~\citep{kwon2023vllm}.
Because the distribution over the vocabulary is only needed for a single token at a time, the memory footprint is independent of sequence length.

At training time, the LLM maximizes the log-likelihood of the next token:
\begin{equation}
{\ell(\hat{\mathbf{x}}) = \sum_{i=1}^N \log P(\hat x_i | \hat x_1, \ldots, \hat x_{i-1})}.
\end{equation}
Due to the structure of most backbones~\citep{transformer2017,gu2022s4,gu2023mamba}, $f(x_1), f(x_1, x_{2}), \ldots, f(x_1, \ldots, x_{N})$ is efficiently computed in parallel.
However, activations for non-linear layers have to be saved for the backward pass, consuming significant memory.
Most LLM training frameworks make use of aggressive activation checkpointing~\citep{chen2016checkpointing}, sharding~\citep{zero2020}, and specialized attention implementations~\citep{flashattention} to keep this memory footprint manageable.

With the aforementioned optimizations, the final (cross-entropy loss) layer of the LLM becomes by far the biggest memory hog. For large vocabularies, the final cross-entropy layer accounts for the majority of the model's memory footprint at training time (\reffig{teaser_regular}).
For example, the log-probabilities materialized by the cross-entropy layer account for $40\%$ of the memory consumption of Phi 3.5 (Mini)~\citep{phi35} ($|V|=\num{32064}$), $65\%$ of the memory consumption of Llama 3 (8B)~\citep{llama3} ($|V|=\num{128000}$), and $89\%$ of the memory consumption of Gemma 2 (2B)~\citep{gemma2} ($|V|=\num{256128}$).
In fact, the log-probabilities of Gemma 2 (2B) for a single sequence $\vec{x}$ with length $N=\num{80000}$ use the entire available memory of an \qty{80}{GB} H100 GPU. (The sequence length is a factor due to the use of teacher forcing for parallelism.)

We show that a reformulation of the training objective leads to an implementation that has negligible memory consumption above what is required to store the loss and the gradient.

\section{Cut Cross-Entropy}

Consider the cross-entropy loss $\ell_i$ over a single prediction of the next token $P(x_i|x_1\ldots x_{i-1})$:
\begin{align*}
\ell_i(\vec{x}) & = \log \softmax_{x_i}\left(\mat{\mC}^\top \imat{\mE}_i\right) = \imat{\mC}^\top_{x_i} \imat{\mE}_i - \log \sum_j \exp\left(\imat{\mC}^\top_j \imat{\mE}_i\right).
\end{align*}
Here the first term is a vector product over $D$-dimensional embeddings $\imat{\mE}_i=f(x_1\ldots x_{i-1})$ and a classifier $\mat{\mC}$.
The second term is a log-sum-exp operation and is independent of the next token $x_i$.
During training, we optimize all next-token predictions $\boldsymbol{\ell} = \left[\ell_1\ldots \ell_N\right]$ jointly using teacher forcing:
\begin{align}
\boldsymbol{\ell} = \left(\mat{\mC}^\top \mat{\mE}\right)_\vec{x} - \log \sum_j \exp(\imat{\mC}_j^\top \mat{\mE} ),\label{eq:ell}
\end{align}
where $\mat{\mE} = \left[\imat{\mE}_1\ldots \imat{\mE}_N\right]$ and $\left( \mat{\mC}^\top\mat{\mE}\right)_\vec{x} = \left[\imat{\mC}^\top_{x_1} \imat{\mE}_1\ldots \imat{\mC}^\top_{x_N} \imat{\mE}_N \right]$.
The first term in \cref{eq:ell} is a combination of an indexing operation and matrix multiplication.
It has efficient forward and backward passes, in terms of both compute and memory, as described in \cref{sec:lin_index}.
The second term in \cref{eq:ell} is a joint log-sum-exp (LSE) and matrix multiplication operation.
\cref{sec:lse-forward} describes how to compute the forward pass of this linear-log-sum-exp operation efficiently using a joint matrix multiplication and reduction kernel.
\cref{sec:lse-backward} describes how to compute its backward pass efficiently by taking advantage of the sparsity of the gradient over a large vocabulary.
Putting all the pieces together yields a memory-efficient low-latency cross-entropy loss.

\begin{figure}[t]
    \captionsetup[subfigure]{justification=centering}
    \begin{subfigure}[t]{.22\linewidth}
      \includegraphics[width=\linewidth,page=4]{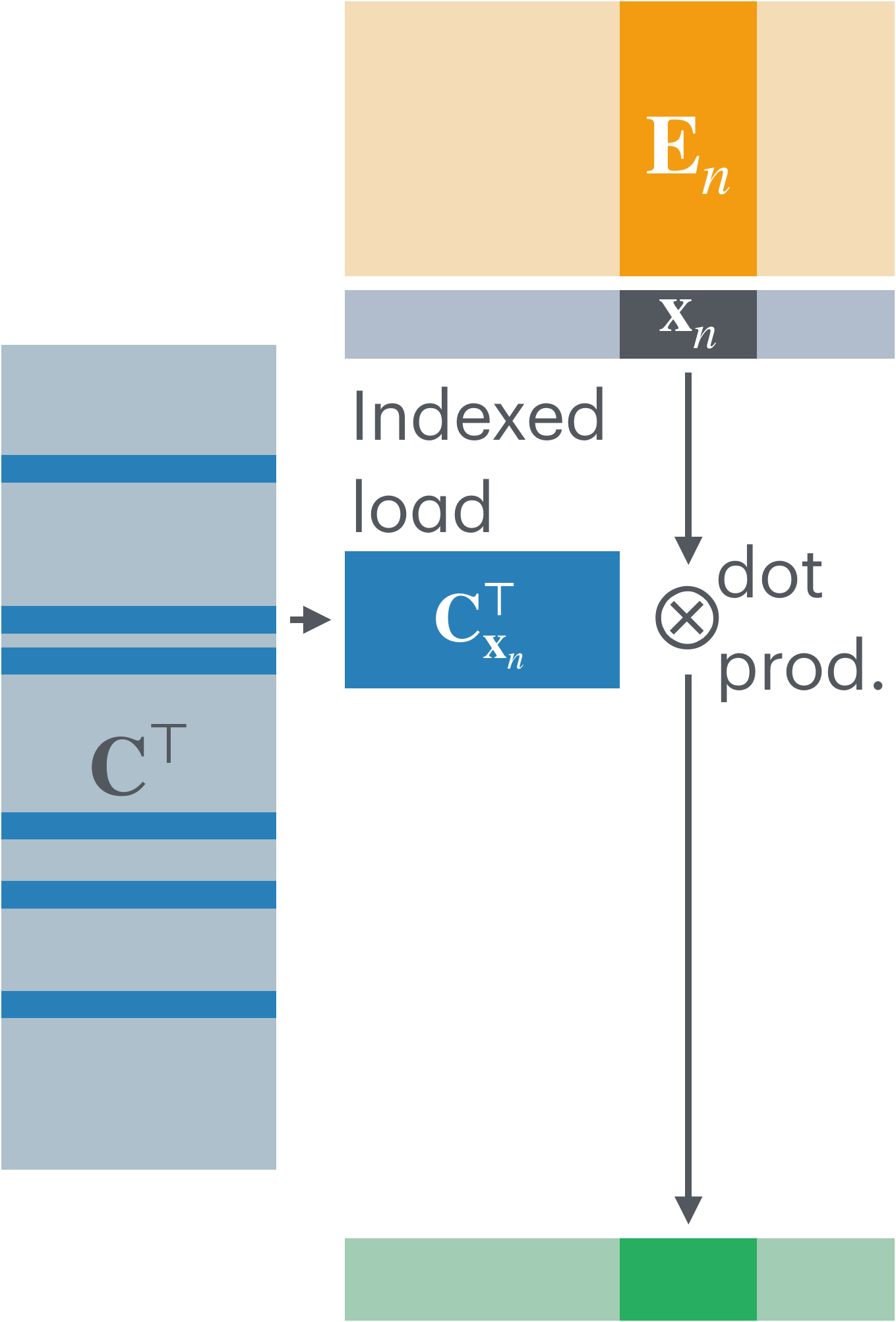}
      \caption{Indexed matmul\\ (forward)}
      \lblfig{imatmul}
    \end{subfigure}\hfill
    \begin{subfigure}[t]{.22\linewidth}
        \includegraphics[width=\linewidth,page=5]{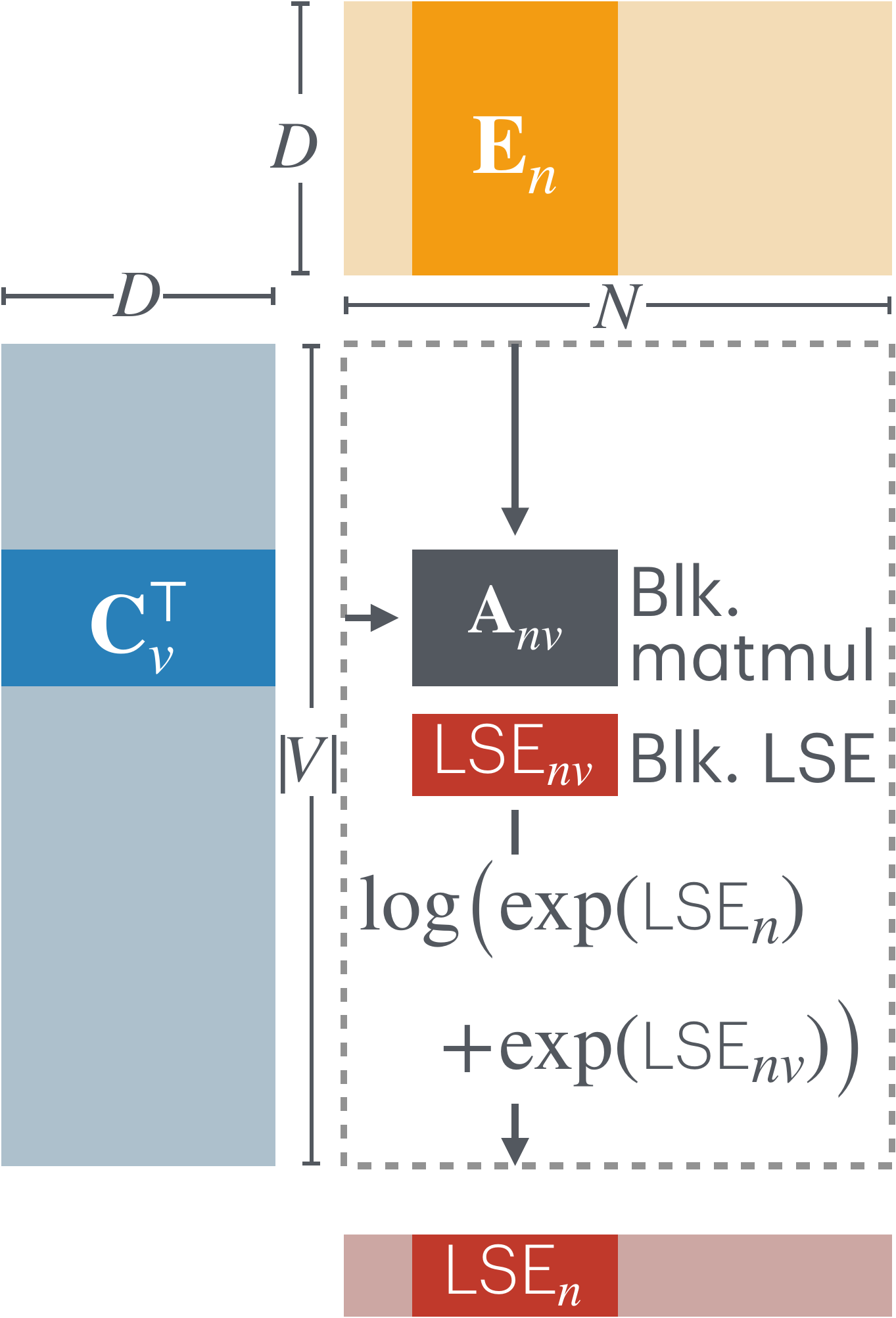}
        \caption{Linear-log-sum-exp,\\forward pass}
        \lblfig{lse_fwd}
      \end{subfigure}\hfill
      \begin{subfigure}[t]{.4634\linewidth}
        \includegraphics[width=\linewidth,page=6]{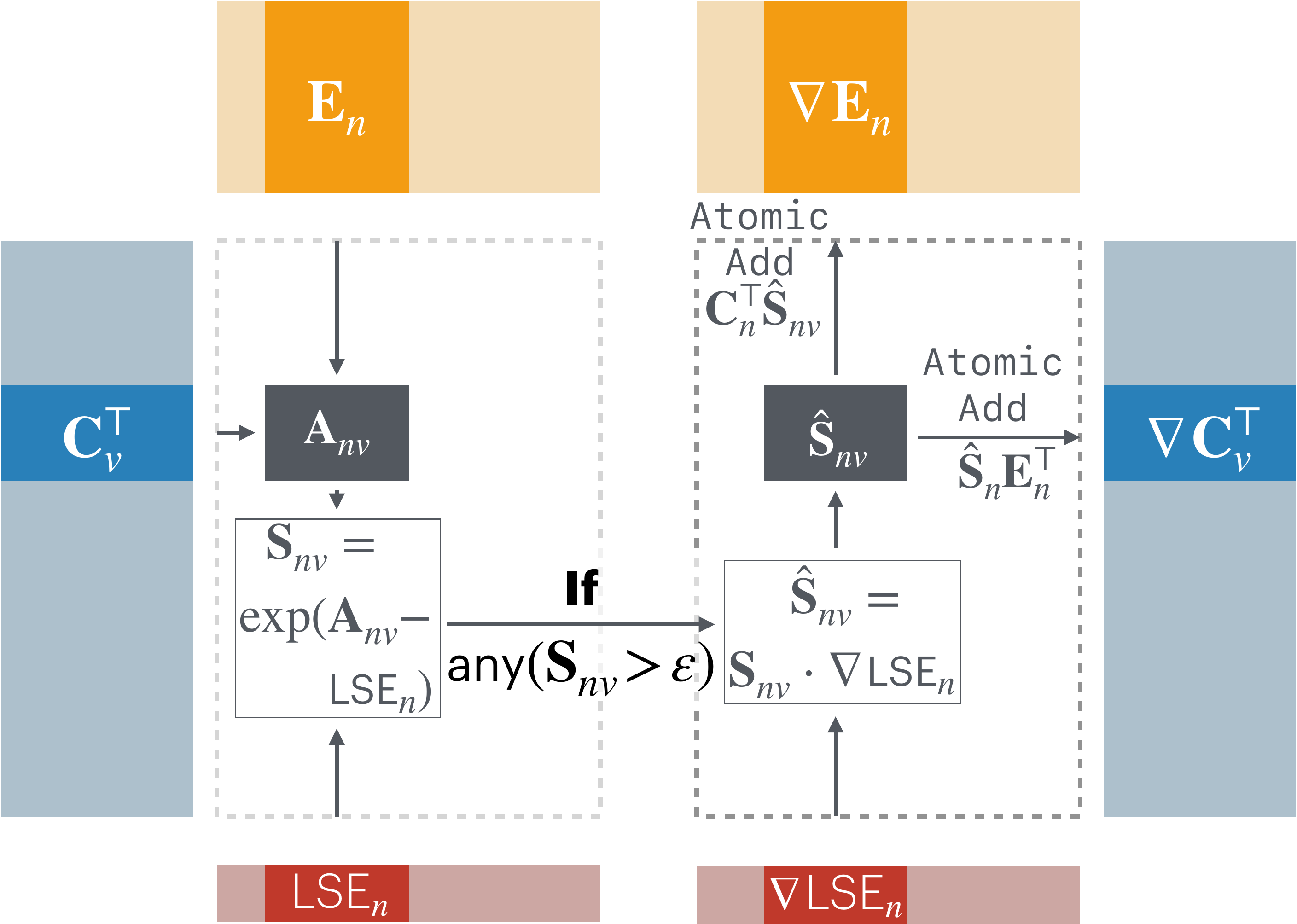}
        \caption{Linear-log-sum-exp,\\backward pass}
        \lblfig{lse_bck}
      \end{subfigure}
        \caption{Access patterns and computation of blockwise (a) indexed matrix multiplication, (b) linear-log-sum-exp forward pass, and (c) linear-log-sum-exp backward pass.
        See \cref{alg:imatmul,alg:lse_fwd,alg:lse_bck} for the corresponding algorithms.}
    \label{fig:lse-forward}
\end{figure}

\subsection{Memory-Efficient Indexed Matrix Multiplication}
\label{sec:lin_index}

A naive computation of indexed matrix multiplication involves either explicit computation of the logits $\mat{\mC}^\top \mat{\mE}$ with an $O(N|V|)$ memory cost, or indexing into the classifier $\mat{\mC}_\vec{x} = \left[\imat{\mC}_{x_1}\ldots \imat{\mC}_{x_N}\right]$ with an $O(ND)$ memory cost.
Our implementation fuses the classifier indexing $\mat{\mC}_\vec{x}$ with the consecutive dot product between columns $\imat{\mC}_{x_i}$ and $\imat{\mE}_i$ in a single CUDA/Triton kernel~\citep{tillet2019triton}.
Our kernel retrieves the value $\ivec{x}_i$, the $\ivec{x}_i$-th column from $\mat{\mC}$, and the $i$-th column from $\mat{\mE}$, and stores them in on-chip shared memory (SRAM).
It then performs a dot product between $\imat{\mC}_{\ivec{x}_i}$ and $\imat{\mE}_i$ and writes the result into global memory.
The kernel uses only on-chip SRAM throughout and does not allocate any GPU memory.
For efficiency, we perform all operations blockwise to make the best use of GPU cache structure.
\refalg{imatmul} and \reffig{imatmul} summarize the computation and access patterns.

\begin{algorithm}[t]
    \begin{tabularx}{\textwidth}{lX}
      \textbf{Inputs:} & $\mat{\mE} \in \mathbb{R}^\dim{D}{N}$, $\mat{\mC} \in \mathbb{R}^\dim{D}{|V|}$, $\vec{x} \in \mathbb{R}^{N}$.\\
      & Block sizes $\BN$ and $\BD$.\\
      \textbf{Outputs:} &  $\vec{o} = (\mat{\mC}^\top\mat{\mE})_\vec{x} \in \mathbb{R}^N$
    \end{tabularx}
    \vspace{0.25em}
    \hrule
    \vspace{0.25em}
    \begin{algorithmic}
        \For{blocks $\mat{\mE}_{n}$, $\vec{x}_n$}\Comment{Divide $\mat{\mE}$ and $\vec{x}$ into blocks of size $\dim{D}{\BN}$ and $\BN$, respectively}
        \State $\vec{o}_{n} = \vec{0}_\BN$\Comment{Zero vector of size $\BN$ in on-chip SRAM}
        \For{blocks $\mat{\mE}_{n,d}$}\Comment{Divide $\mat{\mE}_n$ into blocks of size $\dim{\BD}{\BN}$}
        \State $\mat{c} = \mat{\mC}_{\vec{x}_n,d}$ \Comment{Indexed load into on-chip SRAM}
        \State $\vec{o}_n \pluseq \mat{\mE}_{n,d} \cdot \mat{c}$ \Comment{Column-wide dot product}
        \EndFor
        \State write $\vec{o}_n$\Comment{From on-chip SRAM to main GPU memory}
        \EndFor
    \end{algorithmic}
    \caption{Memory-efficient indexed matrix multiplication}
    \lblalg{imatmul}
\end{algorithm}

\subsection{Memory-efficient Linear-log-sum-exp, Forward Pass}
\label{sec:lse-forward}

Implementing a serial memory-efficient linear-log-sum-exp  is fairly straightforward: use a triple for-loop.
The innermost loop computes the dot product between $\imat{\mC}_v$ and $\imat{\mE}_n$ for the $v$-th token and the $n$-th batch element.
The middle loop iterates over the vocabulary, updating the log-sum-exp (LSE) along the way.
Finally, the outermost loop iterates over all batch elements.
Parallelizing over the outermost loop is trivial and would expose enough work to saturate the CPU due to the number of tokens in training batches (commonly in the thousands).
Parallelization that exposes enough work to saturate the GPU is more challenging.

Let us first examine how efficient matrix multiplication between the batch of model output embeddings $\mat{\mE} \in \mathbb{R}^\dim{D}{N}$ and the classifier $\mat{\mC} \in \mathbb{R}^\dim{D}{|V|}$ is implemented on modern GPUs~\citep{kerr2017cutlass}.
A common method is to first divide the output $\mat{O} = \mat{\mC}^\top\mat{\mE} \in \mathbb{R}^\dim{|V|}{N}$ into a set of blocks of size $\dim{\BV}{\BN}$.
Independent CUDA blocks retrieve the corresponding parts $\mat{\mE}_n$ of $\mat{\mE}$ with size $\dim{D}{\BN}$ and blocks $\mat{\mC}_m$ of $\mat{\mC}$ with size $\dim{D}{\BV}$, and perform the inner product $\mat{O}_{nm} = \mat{\mC}_m^\top  \mat{\mE}_n$ along the $D$ dimension.
Due to limited on-chip SRAM, most implementations use a for-loop for large values of $D$.
They loop over smaller size $\dim{\BD}{\BN}$ and  $\dim{\BD}{\BV}$ blocks and accumulate $\mat{O}_{nv} = \sum_{d} \mat{\mC}_{vd}^\top  \mat{\mE}_{nd}$ in SRAM.
Each CUDA block then writes $\mat{O}_{nm}$ back into global memory.
This method exposes enough work to the GPU and makes efficient use of SRAM and L2 cache.

To produce $\textrm{log-sum-exp}(\mat{\mC}^\top \mat{\mE})$, we use the same blocking and parallelization strategy as matrix multiplication.
Each block first computes a matrix multiplication, then the log-sum-exp along the vocabulary dimension $m$ for its block, and finally updates $\vec{\mLSE}$ with its result.

Note that multiple CUDA blocks are now all writing to the same location of $\vec{\mLSE}$.
This includes blocks in the same input range $n$ but different vocabulary ranges $m$.
We use a spin-lock on an atomic operation in global memory to synchronize the updates by different CUDA blocks as this is simple to implement in our Triton framework and incurs little overhead.
Alternative methods, such as an atomic compare-and-swap loop, may perform better when implementing in CUDA directly.

\refalg{lse_fwd} and \reffig{lse_fwd} summarize the computation and access patterns.

\begin{algorithm}
    \begin{tabularx}{\textwidth}{lX}
      \textbf{Inputs:} & $\mat{\mE} \in \mathbb{R}^\dim{D}{N}$ and $\mat{\mC} \in \mathbb{R}^\dim{D}{|V|}$.\\
      & Block sizes $\BN$, $\BV$, and $\BD$.\\
      \textbf{Outputs:} &  $\vec{\mLSE} = \log\sum_j\exp(\imat{\mC}^\top_j\mat{\mE}) \in \mathbb{R}^N$
    \end{tabularx}
    \vspace{0.25em}
    \hrule
    \vspace{0.25em}
    \begin{algorithmic}
    \State $\vec{\mLSE} = -\vec{\infty}_N$\Comment{$-\infty$ vector of size $N$ in main GPU memory}
    \For{all pairs of blocks $\mat{\mE}_n$, $\mat{\mC}_v$}\Comment{Divide $\mat{\mE}$ and $\mat{\mC}$ into blocks of size $\dim{D}{\BN}$  and $\dim{D}{\BV}$}
        \State $\mat{A}_{nv} = \vec{0}_\dim{\BV}{\BN}$\Comment{Zero matrix of size $\dim{\BV}{\BN}$ in on-chip SRAM}
        \For{blocks $\mat{\mE}_{n,d}$, $\mat{\mC}_{v,d}$}\Comment{Divide $\mat{\mE}_n$ and $\mat{\mC}_v$ into blocks of $\dim{\BD}{\BN}$ and $\dim{\BD}{\BV}$}
            \State $\mat{A}_{nv} \pluseq \mat{\mC}_{v,d}^\top \cdot \mat{\mE}_{n,d}$ \Comment{Blockwise matrix multiplication}
        \EndFor
        \State $\vec{\mLSE}_{nv} = \log \sum \exp(\mat{A}_{nv}^\top)$ \Comment{Numerically stable implementation with max}
        \State $\vec{\mLSE}_{n} = \log(\exp(\vec{\mLSE}_{n}) + \exp(\vec{\mLSE}_{nv})) $\Comment{Locking thread-safe log-add-exp}
    \EndFor
    \end{algorithmic}
    \caption{Memory-efficient linear-log-sum-exp, forward pass}
    \lblalg{lse_fwd}
\end{algorithm}

\subsection{Memory-efficient Linear-log-sum-exp, Backward Pass}
\label{sec:lse-backward}

The backward pass needs to efficiently compute two gradient updates:
\begin{equation*}
\nabla\mat{\mE} = \vec{\lambda}^\top \frac{\partial}{\partial \mat{\mE}}\log\sum\exp(\mat{\mC}^\top\mat{\mE}) \quad\text{and}\quad \nabla\mat{\mC} = \lambda^\top \frac{\partial}{\partial \mat{\mC}}\log\sum\exp(\mat{\mC}^\top\mat{\mE})
\end{equation*}
for a backpropagated gradient $\vec{\lambda} = \nabla\vec{\mLSE}$.
Formally, the gradient is defined as
\begin{equation*}
\nabla\mat{\mE}^\top = \left(\mat{S} \cdot \nabla \vec{\mLSE}\right) \mat{\mC}  \quad\text{and}\quad  \nabla\mat{\mC}^\top = \left(\mat{S} \cdot \nabla \vec{\mLSE}\right)^\top \mat{\mE}\lbleq{lse_grad}
\end{equation*}
where $\mat{S} = \softmax(\mat{\mC}^\top \mat{\mE})$ and $\cdot$ refers to the row-by-row elementwise multiplication of the softmax $\mat{S}$ and the gradient $\nabla \vec{\mLSE}$: $\hat{\mat{S}} = \mat{S} \cdot \nabla \vec{\mLSE}$.

Computationally, the backward pass is a double matrix multiplication $\mat{\mC}^\top \mat{\mE}$ and $\hat{\mat{S}} \mat{\mC}$ or $\hat{\mat{S}}^\top \mat{\mE}$ with intermediate matrices $\mat{S}$ and $\hat{\mat{S}}$ that do not fit into GPU memory and undergo a non-linear operation.
We take a similar approach to the forward pass, recomputing the matrix $\mat{\mC}^\top \mat{\mE}$ implicitly in the GPU's shared memory.
For the backward pass, we do not need to compute the normalization constant of the softmax, since $\mat{S} = \softmax(\mat{\mC}^\top \mat{\mE}) = \exp(\mat{\mC}^\top \mat{\mE} - \vec{\mLSE})$.
This allows us to reuse the global synchronization of the forward pass, and compute $\mat{S}$ efficiently in parallel.

We implement the second matrix multiplication in the main memory of the GPU, as a canonical blockwise implementation would require storing or synchronizing $\mat{S}$.
\refalg{lse_bck} and \reffig{lse_bck} summarize the computation and access patterns.
A naive implementation of this algorithm requires zero additional memory but is slow due to repeated global memory load and store operations.
We use two techniques to improve the memory access pattern: gradient filtering and vocabulary sorting.

\xhdr{Gradient filtering.}
By definition, the softmax $\mat{S}$ sums to one over the vocabulary dimension.
If stored in bfloat16 with a 7-bit fraction, any value below $\varepsilon=2^{-12}$ will likely be ignored due to truncation in the summation or rounding in the normalization.\footnote{The 5 extra bits above the fractional size (7) account for rounding rules, and the consideration that small but not tiny values will likely not get truncated due to the blocking strategies used to compute a sum.}
This has profound implications for the softmax matrix $\mat{S}$: For any column, at most $\frac{1}{\varepsilon} = 4096$ entries have non-trivial values and contribute to the gradient computation.
All other values are either rounded to zero or truncated.
In practice, the sparsity of the softmax matrix $\mat{S}$ is much higher: empirically, in frontier models we evaluate, less than $0.02\%$ of elements are non-zero.
Furthermore, the sparsity of the softmax matrix grows as vocabulary size increases.
In \refalg{lse_bck}, we take advantage of this sparsity and skip gradient computation for any block whose corresponding softmax matrix $S_{nm}$ has only negligible elements.
We chose the threshold $\varepsilon=2^{-12}$ to be the smallest bfloat16 value that is not truncated.
In practice, this leads to a 3.5x speedup without loss of precision in any gradient computation.
See \refsec{analysis} for a detailed analysis.

The efficiency of gradient filtering is directly related to the block-level sparsity of the softmax matrix.
We cannot control the overall sparsity pattern without changing the output.
However, we can change the order of the vocabulary to create denser local blocks for more common tokens.

\begin{algorithm}[t]
    \begin{tabularx}{\textwidth}{lX}
      \textbf{Inputs:} & $\mat{\mE} \in \mathbb{R}^\dim{D}{N}$, $\mat{\mC} \in \mathbb{R}^\dim{D}{|V|}$, $\vec{\mLSE} \in \mathbb{R}^N$, and $\nabla \vec{\mLSE} \in \mathbb{R}^N$.\\
      & Block sizes $\BN$, $\BV$, and $\BD$.\\
      & Accuracy threshold $\varepsilon$.\\
      \textbf{Outputs:} & $\nabla\mat{\mE} \in \mathbb{R}^\dim{D}{N}$, $\nabla\mat{\mC} \in \mathbb{R}^\dim{D}{|V|}$
    \end{tabularx}
    \vspace{0.25em}
    \hrule
    \vspace{0.25em}
    \begin{algorithmic}
    \For{all pairs of blocks $\mat{\mE}_n$, $\mat{\mC}_v$}\Comment{Divide $\mat{\mE}$ and $\mat{\mC}$ into blocks of size $\dim{D}{\BN}$ and $\dim{D}{\BV}$}
        \State $\mat{A}_{nv} = \vec{0}_\dim{\BV}{\BN}$\Comment{Zero matrix of size $\dim{\BV}{\BN}$ in on-chip SRAM}
        \For{blocks $\mat{\mE}_{n,d}$, $\mat{\mC}_{v,d}$}\Comment{Divide $\mat{\mE}_n$ and $\mat{\mC}_v$ into blocks of $\dim{\BD}{\BN}$ and $\dim{\BD}{\BV}$}
            \State $\mat{A}_{nv} \pluseq \mat{\mC}_{v,d}^\top \cdot \mat{\mE}_{n,d}$ \Comment{Blockwise matrix multiplication}
        \EndFor
        \State $\mat{S}_{nv} = \exp(\mat{A}_{nv} - \vec{\mLSE}_n)$ \Comment{Compute the softmax}
        \If{all$(\mat{S}_{nv} < \varepsilon)$}
            \State \textbf{skip} \Comment{Skip computation if below desired numerical precision}
        \EndIf
        \For{blocks $\mat{\mE}_{n,d}$, $\mat{\mC}_{v,d}$}\Comment{Divide $\mat{\mE}_n$ and $\mat{\mC}_m$ into blocks of $\dim{\BD}{\BN}$ and $\dim{\BD}{\BV}$}
            \State $\nabla \mat{\mE}^\top_{n,d} \pluseq \left(\mat{S}_{nv} \cdot \nabla \vec{\mLSE}_n\right) \mat{\mC}_{v,d}$\Comment{Locking thread-safe gradient update}
            \State $\nabla \mat{\mC}^\top_{v,d} \pluseq \left(\mat{S}_{nv} \cdot \nabla \vec{\mLSE}_n\right)^\top \mat{\mE}_{n,d}$\Comment{Locking thread-safe gradient update}
        \EndFor
    \EndFor
    \end{algorithmic}
    \caption{Memory-efficient linear-log-sum-exp, backward pass}
    \lblalg{lse_bck}
\end{algorithm}

\xhdr{Vocabulary sorting.}
Ideally the vocabulary would be ordered such that all tokens with non-trivial gradients would be contiguously located.
This reduces the amount of computation wasted by partially populated blocks -- ideally blocks would either be entirely empty (and thus skipped) or entirely populated.
We heuristically group the non-trivial gradients by ordering the tokens by their average logit. 
Specifically, during the forward pass (described in \cref{sec:lse-forward}) we compute the average logit per token using an atomic addition. 
For the backward pass, we divide the vocabulary dimension $|V|$ into blocks with similar average logit instead of arbitrarily.
This requires a temporary buffer of size $O(|V|)$, about $1$ MB for the largest vocabularies in contemporary LLMs~\citep{gemma2}.

Putting all the pieces together, we arrive at forward and backward implementations of cross-entropy that have a negligible incremental memory footprint without sacrificing speed.
Note that in practice,
we found it to be easier and more memory-efficient to merge the indexed matrix-multiplication backward implementation with the backward pass of the linear-log-sum-exp operator (\refalg{lse_bck}).
The two operations share much of the computation and memory access pattern, see \refalg{cel_bck}.

\section{Analysis}
\lblsec{analysis}

\subsection{Runtime and Memory}

\begin{table}
    \centering
    \setlength{\tabcolsep}{2pt}
    \resizebox{0.975\textwidth}{!}{
    \begin{tabular}{@{\makebox[2.0em][l]{\rownumber}}
        lc rc rc rc rc rc r}
    \toprule
    && \multicolumn{3}{c}{Loss} &&  \multicolumn{3}{c}{Gradient} && \multicolumn{3}{c}{Loss+Gradient} \\
    \cmidrule{3-5}
    \cmidrule{7-9} \cmidrule{11-13}
    Method && Memory && Time  && Memory  && Time && Memory && Time  \\
    \midrule
    Lower bound && \qty{0.004}{MB} && && \qty{1161}{MB} &&  && \qty{1161}{MB} &&  \gdef\rownumber{\stepcounter{magicrownumbers}\arabic{magicrownumbers})} \\
    \midrule
    \ours (Ours) && \textbf{\qty{1}{MB}} && \textbf{\qty{46}{ms}} && \textbf{\qty{1163}{MB}} && \qty{100}{ms} && \textbf{\qty{1164}{MB}} && \qty{145}{ms} \\
    Liger Kernels~\citep{liger2024}\tablefootnote{The gradient and loss are computed simultaneously, not in separate forward/backward passes.}
         && \qty{1474}{MB} && \qty{304}{ms} &&  && && \qty{1474}{MB} && \qty{304}{ms} \\
    \citet{torchtune2024} (8 chunks) && \qty{8000}{MB} && \qty{55}{ms} && \qty{1630}{MB} && \qty{115}{ms} && \qty{9631}{MB} && \qty{169}{ms} \\
    \texttt{torch.compile} && \qty{4000}{MB} && \qty{49}{ms} && \qty{12000}{MB} && \textbf{\qty{92}{ms}} && \qty{16000}{MB} && \textbf{\qty{143}{ms}} \\
    Baseline && \qty{24000}{MB} && \qty{82}{ms} && \qty{16000}{MB} && \qty{122}{ms} && \qty{28000}{MB} && \qty{208}{ms} \\
    \midrule
    \ours (No Vocab Sorting) && \qty{0.09}{MB} && \qty{45}{ms} && \qty{1162}{MB} && \qty{115}{ms} && \qty{1162}{MB} && \qty{159}{ms} \\
    \ours (No Grad. Filter) && \qty{0.09}{MB} && \qty{45}{ms} && \qty{1163}{MB} && \qty{314}{ms} && \qty{1162}{MB} && \qty{357}{ms} \\
    \ourskahan && \qty{1}{MB} && \qty{47}{ms} && \qty{2325}{MB} && \qty{114}{ms} && \qty{2326}{MB} && \qty{160}{ms} \\
    \oursfullc && \qty{1}{MB} && \qty{47}{ms} && \qty{2326}{MB} && \qty{268}{ms} && \qty{2326}{MB} && \qty{313}{ms} \\
    \oursfulle && \qty{1}{MB} && \qty{47}{ms} && \qty{2326}{MB} && \qty{247}{ms} && \qty{2326}{MB} && \qty{292}{ms} \\
    \bottomrule
    \end{tabular}}
    \caption{Peak memory footprint and time to compute the loss, its gradient, and their combination.
    Note that intermediate buffers can often (but not always) be reused between the loss and gradient computation, resulting in lower peak memory consumption than the sum of the parts. Batch of \num{8192} tokens with a vocabulary size of \num{256000} and hidden dimension 2304. Embedding and classifier matrix taken during Gemma 2 (2B) training on Alpaca. Measured on an A100-SXM4 GPU with \qty{80}{GB} of RAM, PyTorch 2.4.1, CUDA 12.4, rounded to closest MB.
    Some numbers are multiples of \num{1000} due to dimensions chosen and PyTorch's allocation strategy.
    `Lower bound' is the amount of memory required for the output buffer(s), \ie $\nabla \mat{\mE}$ and $\nabla \mat{\mC}$, this is the lower bound for the memory footprint of any method. Results averaged over 5 seeds.}
    \label{tab:perf}
    \gdef\rownumber{}
\end{table}

First we examine the runtime and memory of various implementations of the cross-entropy loss $\log \softmax_{x_i}(\mat{\mC}^\top \mat{\mE})$.
We consider a batch of \num{8192} tokens with a vocabulary size of \num{256000} and hidden dimension \num{2304}.
This corresponds to Gemma 2 (2B)~\citep{gemma2}.
We use the Alpaca dataset~\citep{taori2023alpaca} for inputs and labels and Gemma 2 (2B) Instruct weights to compute $\mat{\mE}$ and for $\mat{\mC}$.
The analysis is summarized in \cref{tab:perf}.

The baseline implements the loss directly in PyTorch~\citep{paszke2019pytorch}.
This is the default in popular frameworks such as Torch Tune~\citep{torchtune2024} and Transformers~\citep{hf2019transformers}.
This method has reasonable throughput but a peak memory usage of \qty{28000}{MB} of GPU memory to compute the loss+gradient (\cref{tab:perf} row 5).
Due to memory fragmentation, just computing the loss+gradient for the classifier head requires an \qty{80}{GB} GPU.
\texttt{torch.compile}~\citep{ansel2024pytorch2} is able to reduce memory usage by 43\% and computation time by 33\%, demonstrating the effectiveness of kernel fusion (\cref{tab:perf} row 4 vs.\ 5).
Torch Tune~\citep{torchtune2024} includes a method to compute the cross-entropy loss that divides the computation into chunks and uses \texttt{torch.compile} to save memory.
This reduces memory consumption by 65\% vs.~Baseline and by 40\% vs.~\texttt{torch.compile} (to \qty{9631}{MB}, see \cref{tab:perf} row 3 vs.\ 4 and 5).
Liger Kernels~\citep{liger2024} provide a memory-efficient implementation of the cross-entropy loss that, like Torch Tune, makes uses of chunked computation to reduce peak memory usage.
While very effective at reducing the memory footprint, using 95\% less memory than Baseline, it has a detrimental effect on latency, more than doubling the wall-clock time for the computation (\cref{tab:perf}, row 2 vs.\ 4).
The memory usage of \ours grows with $O(N + |V|)$, as opposed to $O(N \times |V|)$ for Baseline, \texttt{torch.compile}, and Torch Tune, and $O(N \times D)$ for Liger Kernels.
In practice, \ours has a negligible memory footprint regardless of vocabulary size or sequence length.

Compared to the fastest method, \texttt{torch.compile}, \ours computes the loss slightly faster (5\%, 4ms, \cref{tab:perf} row 1 vs.\ 4).
This is because \ours does not write all the logits to global memory.
\ours computes the loss+gradient slightly slower (6\%, \qty{2}{ms}). While \ours needs to recompute $\mat{\mC}^\top \mat{\mE}$, it is able to save time in other parts of the computation.
See \cref{apx:perf-breakdown} for a breakdown of the backwards pass of \ours and Baseline.
This increase is largely negligible as the forward+backward pass for even a small LLM (2B parameters) is on the order of seconds.

\begin{wrapfigure}{R}{0.5\textwidth}
    \centering
    \vspace{-1em}
    \includegraphics[width=0.95\linewidth,page=5]{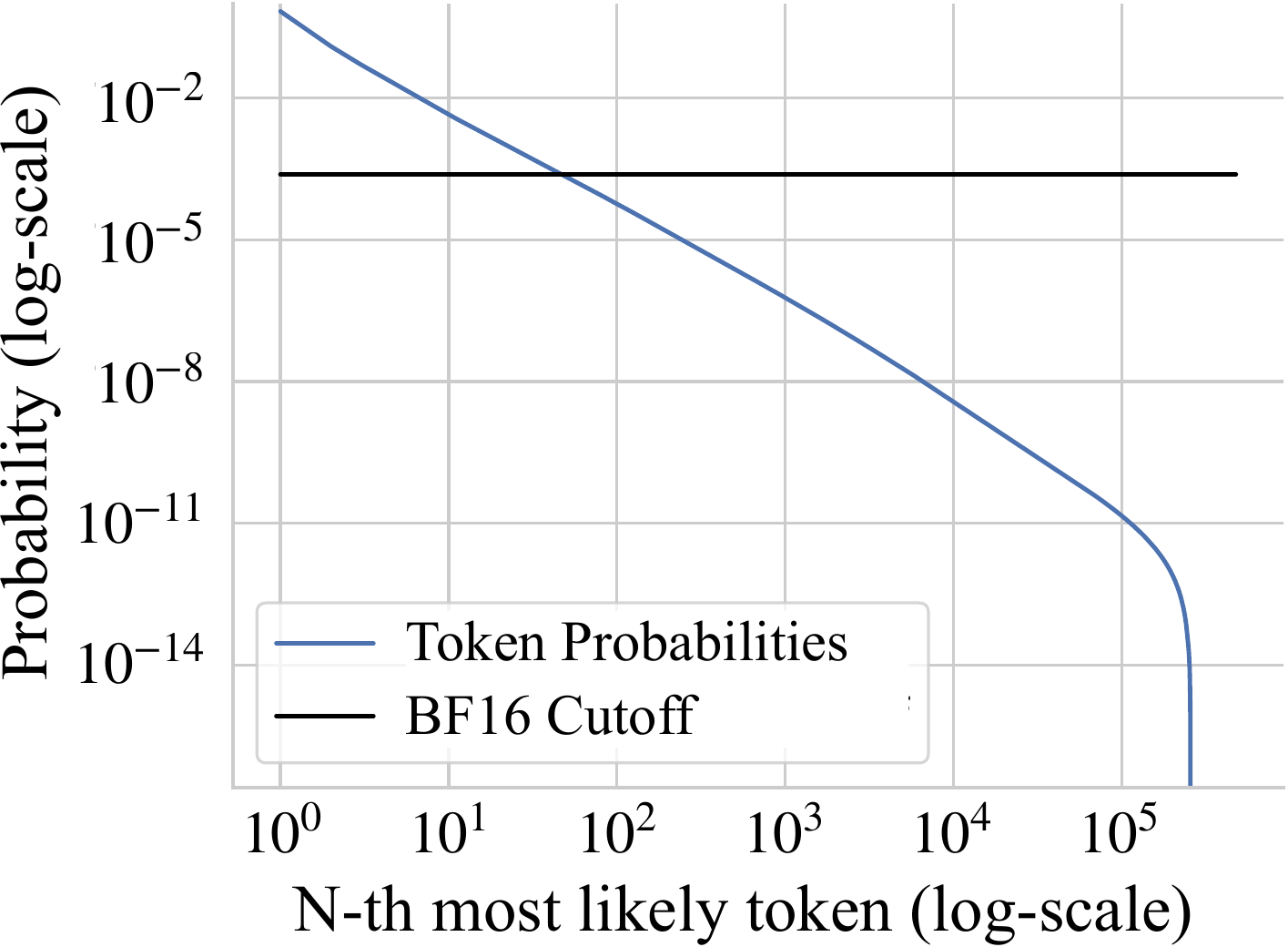}
    \caption{Average probability for the $i$th most likely token, log-log plot. The probabilities very quickly vanish below numerical precision.}
    \lblfig{prop-dist}
    \vspace{-2em}
\end{wrapfigure}

The performance of \ours is enabled several factors.
Without vocabulary sorting \ours takes 15\% (\qty{23}{ms}) longer (\cref{tab:perf} row 1 vs.\ 6) and without gradient filtering it is 3.4x (\qty{356}{ms}) longer (row 1 vs.\ 7). 
\ours utilizes the final gradient floating point type (typically bf16) for summation in global memory.
For increased numerical stability, we experiment with Kahan summation~\citep{kahan1965pracniques} with a higher time and memory cost (\cref{tab:perf} row 1 vs.\ 8). 
We can further incraese the numerical stability by selectively applying gradient filtering to just $\nabla E$ and $\nabla C$.
When combined with Kahan summation, removing gradient filtering from either $\nabla \mC$ or $\nabla \mE$ results in a similar decrease of performance (\cref{tab:perf} row 9 or 10 vs.\ 8).
The last variant (\oursfullc) is particularly interesting for pretraining, where the numerical precision makes a difference.
For fine-tuning all variants of CCE perform equivalently, as shown in \cref{sec:train_stab}.

In \cref{apx:ignored-filter}, we demonstrate that \ours (and other methods) can be made up to 3 times faster by removing tokens that are ignored.
In \cref{apx:additional-results} we benchmark with more models. We find that as the vocabulary size ($|V|$) to hidden size ($D$) ratio decreases, \ours's advantage in computation time for Loss+Gradient decreases, but continues to save a substantial amount of memory.

\subsection{Gradient Filtering}

\cref{fig:prop-dist} shows the sorted softmax probability of vocabulary entries.
Note that the probabilities vanish very quickly and, for the top $10^5$ most likely tokens, there is a linear relationship between $\log \textrm{rank}$ and $\log \textrm{probability}$.
Second, by the $\sim$50th most likely token, the probability has fallen bellow our threshold for gradient filtering.

This explains why we are able to filter so many values from the gradient computation without affecting the result.
At these sparsity levels, most blocks of the softmax matrix $\mat{S}$ are empty.

\begin{figure}
    \captionsetup[subfigure]{justification=centering}
    \begin{subfigure}[t]{.48\linewidth}
        \includegraphics[width=\linewidth,page=4]{figures/training_cropped.pdf}
        \caption{Gemma 2 2B}
    \end{subfigure}\hfill
    \begin{subfigure}[t]{.48\linewidth}
        \includegraphics[width=\linewidth,page=3]{figures/training_cropped.pdf}
        \caption{Phi 3.5 Mini}
    \end{subfigure}\\
    \\
    \\
    \begin{subfigure}[t]{.48\linewidth}
        \includegraphics[width=\linewidth,page=2]{figures/training_cropped.pdf}
        \caption{Qwen 2.5 7B}
    \end{subfigure}\hfill
    \begin{subfigure}[t]{.48\linewidth}
        \includegraphics[width=\linewidth,page=1]{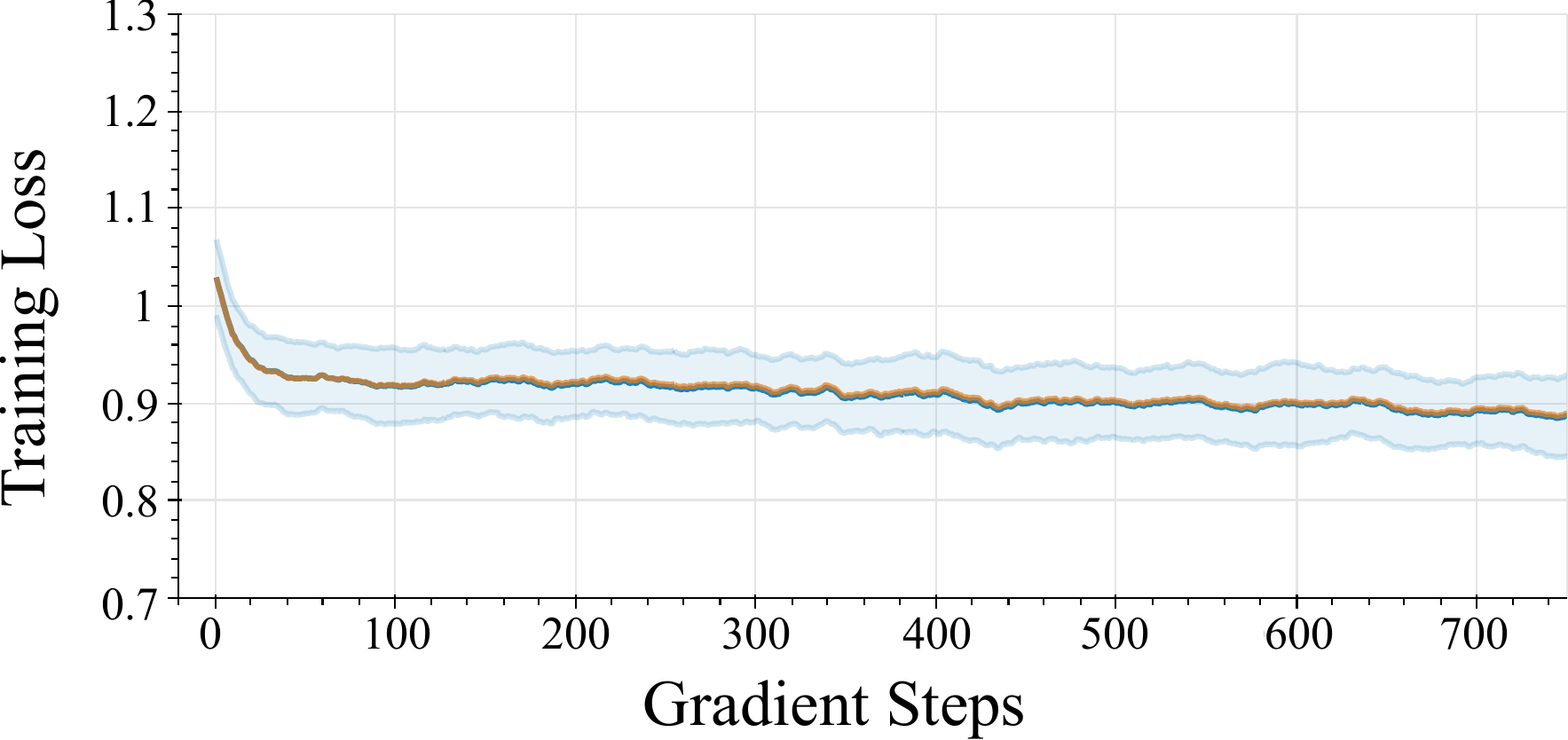}
        \caption{Mistral Nemo}
    \end{subfigure}
    \caption{Training loss curves for four models on the Alpaca dataset~\citep{taori2023alpaca}. The loss curves for \ours and \texttt{torch.compile} are nearly indistinguishable, showing that the gradient filtering in \ours does not impair convergence. Results averaged over 5 seeds.}
    \lblfig{training-curves}
\end{figure}

\subsection{Training Stability}
\label{sec:train_stab}

\xhdr{Fine-tuning.}
We fine-tune Qwen 2.5 7B Instruct~\citep{qwen2.5}, Phi 3.5 Mini Instruct~\citep{phi35}, Gemma 2 2B Instruct~\citep{gemma2}, and Mistral NeMo~\citep{mistralnemo} on the Alpaca Dataset~\citep{taori2023alpaca} using
\ours and \texttt{torch.compile} as the control. \ours and \texttt{torch.compile} have indistinguishable loss curves, demonstrating that the gradient filtering in \ours does not impair convergence (\cref{fig:training-curves}).

\xhdr{Pretraining.}
In our initial experiments using \ours for pretraining, we found that validation perplexity suffered due to two sources of error.
First, gradient filtering when applied to $\nabla C$ causes no gradient to be propagated to tokens that have little to no support in the training set. This does not cause issues when fine-tuning but does when pretraining.  Second, \ours performs a summation in global memory. It is most efficient to perform this reduction in the desired final floating point type. In pretraining, the resulting loss of precision reduces performance. We use Kahan summation~\citep{kahan1965pracniques} to recover this loss of precision.
This changes correspond to \oursfullc.

\begin{figure}
    \captionsetup[subfigure]{justification=centering}
    \begin{subfigure}[t]{.48\linewidth}
        \includegraphics[width=\linewidth,page=1]{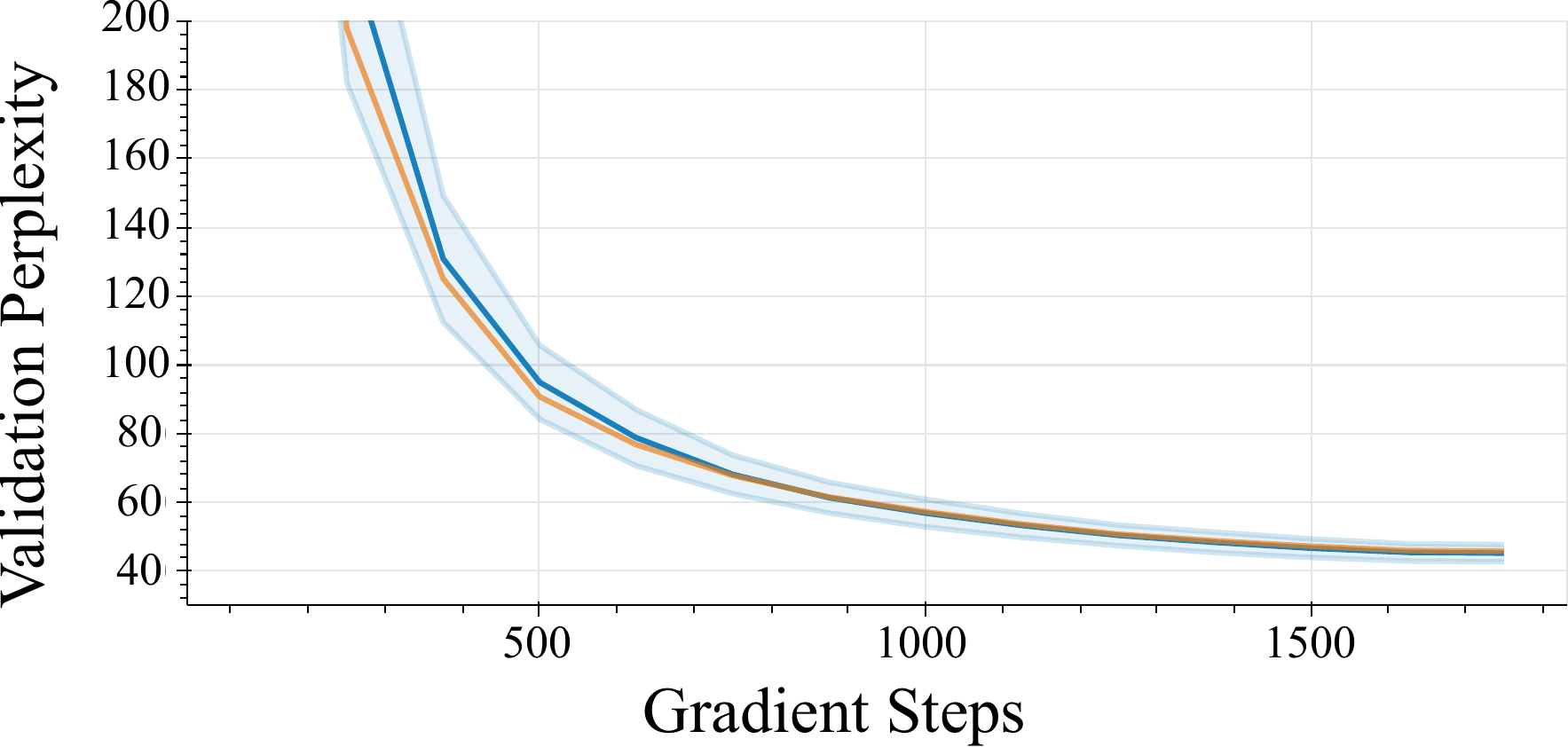}
        \caption{Gemma 2 2B}
    \end{subfigure}\hfill
    \begin{subfigure}[t]{.48\linewidth}
        \includegraphics[width=\linewidth,page=2]{figures/pretrain-plots_cropped.pdf}
        \caption{Phi 3.5 Mini}
    \end{subfigure}\\
    \\
    \\
    \begin{subfigure}[t]{.48\linewidth}
        \includegraphics[width=\linewidth,page=3]{figures/pretrain-plots_cropped.pdf}
        \caption{Qwen 2.5 7B}
    \end{subfigure}\hfill
    \begin{subfigure}[t]{.48\linewidth}
        \includegraphics[width=\linewidth,page=4]{figures/pretrain-plots_cropped.pdf}
        \caption{Mistral Nemo}
    \end{subfigure}
    \caption{Validation perplexity curves for four models on trained using 5\% of the Open WebText dataset~\citep{gokaslan2019openweb}. The validation set is a 0.25\% subset of Open WebText that does not overlap with the train set. We find that \oursfullc matches \texttt{torch.compile}. Results averaged over 5 seeds.}
    \label{fig:pretrain-val-ppl-curves}
\end{figure}

We pretrain Qwen 2.5 7B Instruct~\citep{qwen2.5}, Phi 3.5 Mini Instruct~\citep{phi35}, Gemma 2 2B Instruct~\citep{gemma2}, and Mistral NeMo~\citep{mistralnemo} on the 5\% of the Open WebText Dataset~\citep{gokaslan2019openweb} using \oursfullc and \texttt{torch.compile}. We report validation perplexity on a held-out  0.25\% of Open WebText and find that \oursfullc produces identical curves as \texttt{torch.compile} (\cref{fig:pretrain-val-ppl-curves}).

We make two notes about \oursfullc.  First, the increased memory usage of \oursfullc vs.\ \ours is due to temporary buffers used in the backward pass. The size of these buffers is typically less than the amount of free memory needed to rematerialize activations when using activation/gradient checkpoint~\citep{chen2016checkpointing}. Thus \oursfullc often shares the same memory saving benefits as \ours. Second, the increased computation time of \oursfullc vs.\ \texttt{torch.compile} is often offset by the larger batch sizes \oursfullc enables. In our experiments with Mistral NeMo, \oursfullc enabled doubling the batch size, thereby decreasing training time by 2 hours (16\%) compared to \texttt{torch.compile}.

\section{Discussion}

As vocabulary size $|V|$ has grown in language models, so has the memory footprint of the loss layer.
The memory used by this one layer dominates the training-time memory footprint of many recent language models.
We described \ours, an algorithm to compute \mbox{$\ell_i = \log \softmax_i(\mat{\mC}^T f(x_1 \ldots x_{i-1}))$} and its gradient with negligible memory footprint.

Beyond the immediate impact on compact large-vocabulary LLMs, as illustrated in \reffig{teaser}, we expect that \ours may prove beneficial for training very large models.
Specifically, very large models are trained with techniques such as pipeline parallelism~\citep{huang2019gpipe,narayanan20139pipedream}. Pipeline parallelism works best when all stages are equally balanced in computation load. Achieving this balance is easiest when all blocks in the network have similar memory-to-computation ratios. The classification head is currently an outlier, with a disproportionately high memory-to-computation ratio. \ours may enable better pipeline balancing or reducing the number of stages.

We implemented \ours using Triton~\citep{tillet2019triton}. Triton creates efficient GPU kernels and enables rapid experimentation but has some limitations in control flow. Specifically, the control flow must be specified at the block level and therefore our thread-safe log-add-exp and gradient filtering are constrained to operate at the block level as well. We expect that implementing \ours in CUDA may bring further performance gains because control flow could be performed at finer-grained levels.

It could also be interesting to extend \ours to other classification problems where the number of classes is large, such as image classification and contrastive learning.

\bibliography{main,brody-lit-review}

\begin{thebibliography}{41}
\providecommand{\natexlab}[1]{#1}
\providecommand{\url}[1]{\texttt{#1}}
\expandafter\ifx\csname urlstyle\endcsname\relax
  \providecommand{\doi}[1]{doi: #1}\else
  \providecommand{\doi}{doi: \begingroup \urlstyle{rm}\Url}\fi

\bibitem[Abdin et~al.(2024)Abdin, Jacobs, Awan, Aneja, Awadallah, Awadalla,
  Bach, Bahree, Bakhtiari, Behl, et~al.]{phi35}
Marah~I Abdin, Sam~Ade Jacobs, Ammar~Ahmad Awan, Jyoti Aneja, Ahmed Awadallah,
  Hany Awadalla, Nguyen Bach, Amit Bahree, Arash Bakhtiari, Harkirat~S. Behl,
  et~al.
\newblock Phi-3 technical report: {A} highly capable language model locally on
  your phone, 2024.
\newblock URL \url{https://arxiv.org/abs/2404.14219}.

\bibitem[Ansel et~al.(2024)Ansel, Yang, He, Gimelshein, Jain, Voznesensky, Bao,
  Bell, Berard, Burovski, et~al.]{ansel2024pytorch2}
Jason Ansel, Edward~Z. Yang, Horace He, Natalia Gimelshein, Animesh Jain,
  Michael Voznesensky, Bin Bao, Peter Bell, David Berard, Evgeni Burovski,
  et~al.
\newblock Pytorch 2: Faster machine learning through dynamic python bytecode
  transformation and graph compilation.
\newblock In \emph{{ACM} International Conference on Architectural Support for
  Programming Languages and Operating Systems}, 2024.

\bibitem[Chen et~al.(2016)Chen, Xu, Zhang, and Guestrin]{chen2016checkpointing}
Tianqi Chen, Bing Xu, Chiyuan Zhang, and Carlos Guestrin.
\newblock Training deep nets with sublinear memory cost, 2016.
\newblock URL \url{http://arxiv.org/abs/1604.06174}.

\bibitem[Chen et~al.(2023)Chen, Sarokin, Lee, Tang, Chang, Kulik, and
  Grundmann]{chen2023speed}
Yu{-}Hui Chen, Raman Sarokin, Juhyun Lee, Jiuqiang Tang, Chuo{-}Ling Chang,
  Andrei Kulik, and Matthias Grundmann.
\newblock Speed is all you need: On-device acceleration of large diffusion
  models via {GPU}-aware optimizations.
\newblock In \emph{Conference on Computer Vision and Pattern Recognition,
  Workshops}, 2023.

\bibitem[Choromanski et~al.(2021)Choromanski, Likhosherstov, Dohan, Song, Gane,
  Sarl{\'{o}}s, Hawkins, Davis, Mohiuddin, Kaiser, Belanger, Colwell, and
  Weller]{choromanski2020rethinking}
Krzysztof~Marcin Choromanski, Valerii Likhosherstov, David Dohan, Xingyou Song,
  Andreea Gane, Tam{\'{a}}s Sarl{\'{o}}s, Peter Hawkins, Jared~Quincy Davis,
  Afroz Mohiuddin, Lukasz Kaiser, David~Benjamin Belanger, Lucy~J. Colwell, and
  Adrian Weller.
\newblock Rethinking attention with performers.
\newblock In \emph{International Conference on Learning Representations}, 2021.

\bibitem[Dao et~al.(2022)Dao, Fu, Ermon, Rudra, and R{\'{e}}]{flashattention}
Tri Dao, Daniel~Y. Fu, Stefano Ermon, Atri Rudra, and Christopher R{\'{e}}.
\newblock {FlashAttention}: Fast and memory-efficient exact attention with
  {IO}-awareness.
\newblock In \emph{Neural Information Processing Systems}, 2022.

\bibitem[Dubey et~al.(2024)Dubey, Jauhri, Pandey, Kadian, Al{-}Dahle, Letman,
  Mathur, Schelten, Yang, Fan, et~al.]{llama3}
Abhimanyu Dubey, Abhinav Jauhri, Abhinav Pandey, Abhishek Kadian, Ahmad
  Al{-}Dahle, Aiesha Letman, Akhil Mathur, Alan Schelten, Amy Yang, Angela Fan,
  et~al.
\newblock The {Llama} 3 herd of models, 2024.
\newblock URL \url{https://arxiv.org/abs/2407.21783}.

\bibitem[Gage(1994)]{gage1994bpe}
Philip Gage.
\newblock A new algorithm for data compression.
\newblock \emph{The C Users Journal}, 12\penalty0 (2):\penalty0 23--38, 1994.

\bibitem[Gokaslan et~al.(2019)Gokaslan, Cohen, Pavlick, and
  Tellex]{gokaslan2019openweb}
Aaron Gokaslan, Vanya Cohen, Ellie Pavlick, and Stefanie Tellex.
\newblock Openwebtext corpus, 2019.
\newblock URL \url{http://Skylion007.github.io/OpenWebTextCorpus}.

\bibitem[Goyal et~al.(2017)Goyal, Doll{\'{a}}r, Girshick, Noordhuis,
  Wesolowski, Kyrola, Tulloch, Jia, and He]{goyal2017large}
Priya Goyal, Piotr Doll{\'{a}}r, Ross~B. Girshick, Pieter Noordhuis, Lukasz
  Wesolowski, Aapo Kyrola, Andrew Tulloch, Yangqing Jia, and Kaiming He.
\newblock Accurate, large minibatch {SGD}: Training {ImageNet} in 1 hour, 2017.
\newblock URL \url{http://arxiv.org/abs/1706.02677}.

\bibitem[Grave et~al.(2017)Grave, Joulin, Ciss{\'{e}}, Grangier, and
  J{\'{e}}gou]{joulin2017efficient}
Edouard Grave, Armand Joulin, Moustapha Ciss{\'{e}}, David Grangier, and
  Herv{\'{e}} J{\'{e}}gou.
\newblock Efficient softmax approximation for gpus.
\newblock In \emph{International Conference on Machine Learning}, 2017.

\bibitem[Gu \& Dao(2023)Gu and Dao]{gu2023mamba}
Albert Gu and Tri Dao.
\newblock Mamba: Linear-time sequence modeling with selective state spaces,
  2023.
\newblock URL \url{https://arxiv.org/abs/2312.00752}.

\bibitem[Gu et~al.(2022)Gu, Goel, and R{\'{e}}]{gu2022s4}
Albert Gu, Karan Goel, and Christopher R{\'{e}}.
\newblock Efficiently modeling long sequences with structured state spaces.
\newblock In \emph{International Conference on Learning Representations}, 2022.

\bibitem[Hillis \& Steele(1986)Hillis and Steele]{hillis1986data}
W.~Daniel Hillis and Guy~L. Steele.
\newblock Data parallel algorithms.
\newblock \emph{Commun. {ACM}}, 29\penalty0 (12):\penalty0 1170--1183, 1986.

\bibitem[Hsu et~al.(2024)Hsu, Dai, Kothapalli, Song, Tang, and Zhu]{liger2024}
Pin-Lun Hsu, Yun Dai, Vignesh Kothapalli, Qingquan Song, Shao Tang, and Siyu
  Zhu.
\newblock {Liger-Kernel}: Efficient {Triton} kernels for {LLM} training, 2024.
\newblock URL \url{https://github.com/linkedin/Liger-Kernel}.

\bibitem[Huang et~al.(2019)Huang, Cheng, Bapna, Firat, Chen, Chen, Lee, Ngiam,
  Le, Wu, and Chen]{huang2019gpipe}
Yanping Huang, Youlong Cheng, Ankur Bapna, Orhan Firat, Dehao Chen, Mia~Xu
  Chen, HyoukJoong Lee, Jiquan Ngiam, Quoc~V. Le, Yonghui Wu, and Zhifeng Chen.
\newblock {GPipe}: Efficient training of giant neural networks using pipeline
  parallelism.
\newblock In \emph{Neural Information Processing Systems}, 2019.

\bibitem[Jacobs et~al.(2023)Jacobs, Tanaka, Zhang, Zhang, Song, Rajbhandari,
  and He]{deepspeed-ulysses}
Sam~Ade Jacobs, Masahiro Tanaka, Chengming Zhang, Minjia Zhang, Shuaiwen~Leon
  Song, Samyam Rajbhandari, and Yuxiong He.
\newblock Deepspeed ulysses: System optimizations for enabling training of
  extreme long sequence transformer models, 2023.
\newblock URL \url{https://doi.org/10.48550/arXiv.2309.14509}.

\bibitem[Kahan(1965)]{kahan1965pracniques}
William Kahan.
\newblock Pracniques: further remarks on reducing truncation errors.
\newblock \emph{Communications of the ACM}, 1965.

\bibitem[Kerr et~al.(2017)Kerr, Merrill, Demouth, and Tran]{kerr2017cutlass}
Andrew Kerr, Duane Merrill, Julien Demouth, and John Tran.
\newblock {CUTLASS}: Fast linear algebra in {CUDA C++}, 2017.
\newblock URL
  \url{https://developer.nvidia.com/blog/cutlass-linear-algebra-cuda/}.

\bibitem[Kingma \& Ba(2015)Kingma and Ba]{kingma2015adam}
Diederik~P. Kingma and Jimmy Ba.
\newblock Adam: {A} method for stochastic optimization.
\newblock In \emph{International Conference on Learning Representations}, 2015.

\bibitem[Kitaev et~al.(2020)Kitaev, Kaiser, and Levskaya]{kitaev2020reformer}
Nikita Kitaev, Lukasz Kaiser, and Anselm Levskaya.
\newblock Reformer: The efficient transformer.
\newblock In \emph{International Conference on Learning Representations}, 2020.

\bibitem[Kwon et~al.(2023)Kwon, Li, Zhuang, Sheng, Zheng, Yu, Gonzalez, Zhang,
  and Stoica]{kwon2023vllm}
Woosuk Kwon, Zhuohan Li, Siyuan Zhuang, Ying Sheng, Lianmin Zheng, Cody~Hao Yu,
  Joseph Gonzalez, Hao Zhang, and Ion Stoica.
\newblock Efficient memory management for large language model serving with
  pagedattention.
\newblock In \emph{Symposium on Operating Systems Principles}, 2023.

\bibitem[Li et~al.(2023)Li, Xue, Baranwal, Li, and You]{li2023seqparallel}
Shenggui Li, Fuzhao Xue, Chaitanya Baranwal, Yongbin Li, and Yang You.
\newblock Sequence parallelism: Long sequence training from system perspective.
\newblock In \emph{Association for Computational}, 2023.

\bibitem[Loshchilov \& Hutter(2019)Loshchilov and Hutter]{loshchilov2019adamw}
Ilya Loshchilov and Frank Hutter.
\newblock Decoupled weight decay regularization.
\newblock In \emph{International Conference on Learning Representations}, 2019.

\bibitem[Milakov \& Gimelshein(2018)Milakov and Gimelshein]{milakov2018softmax}
Maxim Milakov and Natalia Gimelshein.
\newblock Online normalizer calculation for softmax, 2018.
\newblock URL \url{http://arxiv.org/abs/1805.02867}.

\bibitem[{Mistral AI Team}(2024)]{mistralnemo}
{Mistral AI Team}.
\newblock Mistral {N}e{M}o, 2024.
\newblock URL \url{https://mistral.ai/news/mistral-nemo/}.

\bibitem[Narayanan et~al.(2019)Narayanan, Harlap, Phanishayee, Seshadri,
  Devanur, Ganger, Gibbons, and Zaharia]{narayanan20139pipedream}
Deepak Narayanan, Aaron Harlap, Amar Phanishayee, Vivek Seshadri, Nikhil~R.
  Devanur, Gregory~R. Ganger, Phillip~B. Gibbons, and Matei Zaharia.
\newblock Pipedream: Generalized pipeline parallelism for {DNN} training.
\newblock In \emph{{ACM} Symposium on Operating Systems Principles}, 2019.

\bibitem[Paszke et~al.(2019)Paszke, Gross, Massa, Lerer, Bradbury, Chanan,
  Killeen, Lin, Gimelshein, Antiga, et~al.]{paszke2019pytorch}
Adam Paszke, Sam Gross, Francisco Massa, Adam Lerer, James Bradbury, Gregory
  Chanan, Trevor Killeen, Zeming Lin, Natalia Gimelshein, Luca Antiga, et~al.
\newblock {PyTorch}: An imperative style, high-performance deep learning
  library.
\newblock In \emph{Neural Information Processing Systems}, 2019.

\bibitem[{Qwen Team}(2024)]{qwen2.5}
{Qwen Team}.
\newblock Qwen2.5: A party of foundation models, September 2024.
\newblock URL \url{https://qwenlm.github.io/blog/qwen2.5/}.

\bibitem[Rabe \& Staats(2021)Rabe and Staats]{rabe2021memeff}
Markus~N. Rabe and Charles Staats.
\newblock Self-attention does not need {O(n\({}^{\mbox{2}}\))} memory, 2021.
\newblock URL \url{https://arxiv.org/abs/2112.05682}.

\bibitem[Rajbhandari et~al.(2020)Rajbhandari, Rasley, Ruwase, and He]{zero2020}
Samyam Rajbhandari, Jeff Rasley, Olatunji Ruwase, and Yuxiong He.
\newblock {ZeRO}: Memory optimizations toward training trillion parameter
  models.
\newblock In \emph{International Conference for High Performance Computing,
  Networking, Storage and Analysis}, 2020.

\bibitem[Rivi{\`{e}}re et~al.(2024)Rivi{\`{e}}re, Pathak, Sessa, Hardin,
  Bhupatiraju, Hussenot, Mesnard, Shahriari, Ram{\'{e}}, Ferret,
  et~al.]{gemma2}
Morgane Rivi{\`{e}}re, Shreya Pathak, Pier~Giuseppe Sessa, Cassidy Hardin,
  Surya Bhupatiraju, L{\'{e}}onard Hussenot, Thomas Mesnard, Bobak Shahriari,
  Alexandre Ram{\'{e}}, Johan Ferret, et~al.
\newblock Gemma 2: Improving open language models at a practical size, 2024.
\newblock URL \url{https://arxiv.org/abs/2408.00118}.

\bibitem[Shoeybi et~al.(2019)Shoeybi, Patwary, Puri, LeGresley, Casper, and
  Catanzaro]{megatronlm}
Mohammad Shoeybi, Mostofa Patwary, Raul Puri, Patrick LeGresley, Jared Casper,
  and Bryan Catanzaro.
\newblock {Megatron-LM}: Training multi-billion parameter language models using
  model parallelism, 2019.
\newblock URL \url{http://arxiv.org/abs/1909.08053}.

\bibitem[Tao et~al.(2024)Tao, Liu, Dou, Muennighoff, Wan, Luo, Lin, and
  Wong]{tao2024scalinglawsvocabularylarger}
Chaofan Tao, Qian Liu, Longxu Dou, Niklas Muennighoff, Zhongwei Wan, Ping Luo,
  Min Lin, and Ngai Wong.
\newblock Scaling laws with vocabulary: Larger models deserve larger
  vocabularies, 2024.
\newblock URL \url{https://arxiv.org/abs/2407.13623}.

\bibitem[Taori et~al.(2023)Taori, Gulrajani, Zhang, Dubois, Li, Guestrin,
  Liang, and Hashimoto]{taori2023alpaca}
Rohan Taori, Ishaan Gulrajani, Tianyi Zhang, Yann Dubois, Xuechen Li, Carlos
  Guestrin, Percy Liang, and Tatsunori~B. Hashimoto.
\newblock Stanford {Alpaca}: An instruction-following {LLaMA} model, 2023.
\newblock URL \url{https://github.com/tatsu-lab/stanford_alpaca}.

\bibitem[Tillet et~al.(2019)Tillet, Kung, and Cox]{tillet2019triton}
Philippe Tillet, Hsiang{-}Tsung Kung, and David~D. Cox.
\newblock Triton: An intermediate language and compiler for tiled neural
  network computations.
\newblock In \emph{{ACM} {SIGPLAN} International Workshop on Machine Learning
  and Programming Languages}, 2019.

\bibitem[{Torch Tune Team}(2024)]{torchtune2024}
{Torch Tune Team}.
\newblock torchtune, 2024.
\newblock URL \url{https://github.com/pytorch/torchtune}.

\bibitem[Vaswani et~al.(2017)Vaswani, Shazeer, Parmar, Uszkoreit, Jones, Gomez,
  Kaiser, and Polosukhin]{transformer2017}
Ashish Vaswani, Noam Shazeer, Niki Parmar, Jakob Uszkoreit, Llion Jones,
  Aidan~N. Gomez, Lukasz Kaiser, and Illia Polosukhin.
\newblock Attention is all you need.
\newblock In \emph{Neural Information Processing Systems}, 2017.

\bibitem[Wang et~al.(2020)Wang, Li, Khabsa, Fang, and
  Ma]{wang2020linformerselfattentionlinearcomplexity}
Sinong Wang, Belinda~Z. Li, Madian Khabsa, Han Fang, and Hao Ma.
\newblock Linformer: Self-attention with linear complexity, 2020.
\newblock URL \url{https://arxiv.org/abs/2006.04768}.

\bibitem[Wolf et~al.(2019)Wolf, Debut, Sanh, Chaumond, Delangue, Moi, Cistac,
  Rault, Louf, Funtowicz, and Brew]{hf2019transformers}
Thomas Wolf, Lysandre Debut, Victor Sanh, Julien Chaumond, Clement Delangue,
  Anthony Moi, Pierric Cistac, Tim Rault, R{\'{e}}mi Louf, Morgan Funtowicz,
  and Jamie Brew.
\newblock Huggingface's transformers: State-of-the-art natural language
  processing, 2019.

\bibitem[Yu et~al.(2023)Yu, Simig, Flaherty, Aghajanyan, Zettlemoyer, and
  Lewis]{yu2023megabyte}
Lili Yu, Daniel Simig, Colin Flaherty, Armen Aghajanyan, Luke Zettlemoyer, and
  Mike Lewis.
\newblock {MEGABYTE}: Predicting million-byte sequences with multiscale
  transformers.
\newblock In \emph{Neural Information Processing Systems}, 2023.

\end{thebibliography}
\bibliographystyle{iclr2025_conference}

\clearpage

\appendix
\setcounter{table}{0}
\setcounter{figure}{0}
\renewcommand\thefigure{A\arabic{figure}}
\renewcommand\thetable{A\arabic{table}}

\section{Notation}

Throughout the paper, we use the following notation conventions. Matrices are bold, capital letters, \eg $\mat{A}$. Indexed matrices are capital letters and are indexed by column and then, optionally, row. For example, given $\mat{A} \in \mathbb{R}^{\dim{N}{M}}$, then \eg $\imat{A}_j$ is the length $N$ vector that is the $j$th column for A, $\imat{A}_{j,i}$ is then the $i$th value in the vector  $\imat{A}_j$. When we combine indexing and transposing, we always index and then transpose.

Vectors are bold lower-case letters, \eg $\vec{x}$, with the exception of $\vec{\mLSE}$ which is the vector containing the log-sum-exp (LSE). Indexed vectors are lower-case letters, $\ivec{x}_i$.

In addition to scalar indexing, we also block index matrices when describing how our algorithms are implemented. In these cases, the matrix and vector will maintain their bold to indicate that the indexing refers to a block and thus are still a matrix or vector.

\begin{table}[h!]
    \centering
    \resizebox{0.95\textwidth}{!}{
    \begin{tabular}{cl}
    \toprule
    Notation & Description \\
    \midrule
    $\mat{\mE}$ & A $\dim{D}{N}$ matrix containing batch of inputs. \\
    $\imat{\mE}_i$ & A $D$-dimensional vector containing the embedding for the $i$th input. \\
    $\mat{\mC}$ & A $\dim{D}{\V}$ classifier matrix used to compute the logit for each token. \\
    $\imat{\mC}_i$ & A $D$-dimensional vector used to create the logit for the $i$th token. \\
    $\vec{x}$ & A length $N$ vector containing the inputs. \\
    $\ivec{x}_i$ & A scalar that is the $i$th input. \\
    $\mat{\mC}_{\ivec{x}_i}$ & A length $D$ containing the vector used to create the logit for the $\ivec{x}_i$th token. \\
    $\mat{\mC}^\top \mat{\mE}$ & A $\dim{\V}{N}$ matrix containing the logits over the vocabulary for each input. \\
    $\left(\mat{\mC}^\top \mat{\mE}\right)_\vec{x}$ & A length $N$ vector where the $ith$ entry is the logit for the $\ivec{x}_i$th token. \\
    $\vec{\mLSE}$ & A length $N$ vector containing the log-sum-exp (LSE) for each input over the vocabulary. \\
    $\mat{\mE}_n$ & The $n$th $\dim{D}{\BN}$ block of $\mat{\mE}$. \\
    $\mat{\mE}_{n,d}$ & The $d$th $\dim{\BD}{\BN}$ block of $\mat{\mE}_n$. \\
    $\left[\left[\vec{a} = \vec{b}^\top\right]\right]$ & An indicator matrix where the value at the $i$th column and $j$th row is 1 if $\ivec{a}_j = \ivec{b}_i$ and 0 otherwise.\\
    \bottomrule
    \end{tabular}}
\end{table}

\section{Removing Ignored Tokens}
\label{apx:ignored-filter}

\begin{table}[t]
    \centering
    \setlength{\tabcolsep}{2pt}
    \resizebox{0.975\textwidth}{!}{
    \begin{tabular}{@{\makebox[2.0em][l]{\rownumber}}
        lc rc rc rc rc rc r}
    \toprule
    && \multicolumn{3}{c}{Loss} &&  \multicolumn{3}{c}{Gradient} && \multicolumn{3}{c}{Loss+Gradient} \\
    \cmidrule{3-5}
    \cmidrule{7-9} \cmidrule{11-13}
    Method && Memory && Time  && Memory  && Time && Memory && Time  \\
    \midrule
    Lower bound && \qty{0.004}{MB} && && \qty{1161}{MB} &&  && \qty{1161}{MB} &&  \gdef\rownumber{\stepcounter{magicrownumbers}\arabic{magicrownumbers})} \\
    \midrule
    \ours (Ours) && \textbf{\qty{245}{MB}} && \textbf{\qty{17}{ms}} && \textbf{\qty{1163}{MB}} && \qty{37}{ms} && \textbf{\qty{1164}{MB}} && \qty{54}{ms} \\
    Liger Kernels~\citep{liger2024}\tablefootnote{The gradient and loss are computed simultaneously, not in separate forward/backward passes.}
         && \qty{1316}{MB} && \qty{301}{ms} &&  && && \qty{1314}{MB} && \qty{303}{ms} \\
    \citet{torchtune2024} (8 chunks) && \qty{3688}{MB} && \qty{23}{ms} && \qty{2789}{MB} && \qty{54}{ms} && \qty{6157}{MB} && \qty{77}{ms} \\
    \texttt{torch.compile} && \qty{1847}{MB} && \qty{19}{ms} && \qty{5490}{MB} && \textbf{\qty{34}{ms}} && \qty{7337}{MB} && \textbf{\qty{53}{ms}} \\
    Baseline && \qty{10997}{MB} && \qty{30}{ms} && \qty{7320}{MB} && \qty{44}{ms} && \qty{12826}{MB} && \qty{75}{ms} \\
    \midrule
    \ours (No Vocab Sorting) && \qty{0.06}{MB} && \qty{17}{ms} && \qty{1162}{MB} && \qty{43}{ms} && \qty{1163}{MB} && \qty{60}{ms} \\
    \ours (No Grad. Filter) && \qty{0.06}{MB} && \qty{17}{ms} && \qty{1163}{MB} && \qty{110}{ms} && \qty{1163}{MB} && \qty{126}{ms} \\
    \ourskahan && \qty{1}{MB} && \qty{18}{ms} && \qty{2325}{MB} && \qty{42}{ms} && \qty{2327}{MB} && \qty{59}{ms} \\
    \oursfullc && \qty{1}{MB} && \qty{18}{ms} && \qty{2326}{MB} && \qty{98}{ms} && \qty{2327}{MB} && \qty{114}{ms} \\
    \oursfulle && \qty{1}{MB} && \qty{18}{ms} && \qty{2325}{MB} && \qty{92}{ms} && \qty{2327}{MB} && \qty{109}{ms} \\
    \bottomrule
    \end{tabular}}
    \caption{\cref{tab:perf} where all methods include a filter that removes tokens that are ignored in loss computation. This simple change represents large improvements in practice. Results averaged over 5 seeds.}
    \label{tab:perf-ignored-filter}
\end{table}

It is common to have tokens that have no loss computation when training LLMs in practice. Examples include padding, the system prompt, user input, \etc. While these tokens must be processed by the backbone -- to enable efficient batching in the case of padding or to give the model the correct context for its prediction in the case of system prompts and use inputs -- they do not contribute directly to the loss.

In all implementations we are aware of, the logits and loss for these ignored tokens is first computed and then set to zero. We notice that this is unnecessary. These tokens can be removed \emph{before} logits+loss computation with no change to the loss/gradient and save a significant amount of computation.

\cref{tab:perf-ignored-filter} shows the performance of all methods in \cref{tab:perf} with a filter that removes ignored tokens before logits+loss computation. This represents a significant speed up for all methods but Liger Kernels. Due to heavy chunking in Liger Kernels to save memory, it is bound by kernel launch overhead, not computation, and therefore reducing the amount of computation does not increase speed. Filtering ignored tokens is also a significant memory saving for most all but \ours (because \ours already uses the minimum amount of memory possible).

\section{Additional Results}
\label{apx:additional-results}

\subsection{Further Performance Analysis}
\label{apx:perf-breakdown}

\cref{tab:perf-breakdown} shows a breakdown of the time spent for different components of in the backward pass of \ours and Baseline. For \ours, we selectively disabled/enabled portions of the kernel and measured the time saved to determine the amount of time taken by that component. For Baseline, we manually implemented each operation of the backward pass and timed them seperately.

\ours spends considerably less time on the cross-entropy loss and softcap portions of the gradient computation. For Baseline, these are very memory intensive operations as there is relatively very little computation done compared the amount of reading/writing. For \ours, the logits are already in SRAM (they were just recomputed) and \ours does not write the result of this computation to main memory, saving a significant amount of time.

Coincidentally, \ours spends a very similar amount of time computing the gradient \wrt the embeddings. \ours spends less time computing the gradient \wrt the classifier. This is because the axis we reduce along for the classifier, N, is shorter than the axis for the embeddings, |V|, and thus leads to less contention on global memory.

Compared to Baseline, \ours saves \qty{30}{ms} on the gradient of the logits \wrt cross-entropy loss, \qty{12}{ms} on the gradient \wrt softcapping, \qty{5}{ms} on the gradient \wrt E, and \qty{15}{ms} on the gradient \wrt C. This saving of \qty{62}{ms} more than offsets the \qty{45}{ms} spent re-computing and applying the gradient filter.

\begin{table}
    \setlength{\tabcolsep}{4pt}
    \centering
    \begin{tabular}{l r r}
    \toprule
    Component & Baseline & \ours \\
    \midrule
    $\mathrm{logits} = \mathrm{softcap}\left(\mat{\mC}^\top \mat{\mE}\right)$ recomputation &  & \qty{45}{ms} (\qty{43.2}{\percent}) \\[0.5em]
    $\nabla \log \softmax_\vec{x}\left(\mathrm{logits}\right)$  & \qty{35}{ms} (\qty{28.5}{\percent}) & \qty{4.7}{ms} (\qty{4.4}{\percent}) \\[0.5em]
    Gradient Filter & & \qty{1.3}{ms} (\qty{1.2}{\percent}) \\[0.5em]
    $\nabla  \mathrm{softcap}\left(\mat{\mC}^\top \mat{\mE}\right)$ & \qty{17}{ms} (\qty{13.7}{\percent}) & \qty{4.7}{ms} (\qty{4.4}{\percent}) \\[0.5em]
    $\nabla \mat{\mE}$ & \qty{37}{ms} (\qty{30.0}{\percent}) & \qty{31}{ms} (\qty{29.6}{\percent}) \\[0.5em]
    $\nabla \mat{\mC}$ & \qty{34}{ms} (\qty{27.7}{\percent}) & \qty{18}{ms} (\qty{17.3}{\percent}) \\
    \bottomrule
    \end{tabular}
    \caption{Performance breakdown for the backward pass of \ours and Baseline. Gemma 2 (\qty{2}{B}) model. Batch of 8192 tokens. Alpaca dataset used to generate inputs.}
    \label{tab:perf-breakdown}
\end{table}

\subsection{Additional Runtime and Memory}
\cref{tab:additional-results} shows additional results for Gemma 2 (\qty{9}{B}), Gemma 2 (\qty{27}{B}), Qwen 2.5 (\qty{7}{B})~\citep{qwen2.5},  Qwen 2.5 (\qty{32}{B}), PHI 3.5 Mini~\citep{phi35}, and Mistral NeMo~\citep{mistralnemo} in the same setting as \cref{tab:perf}.
For each model \ours is able to reduce the total memory consumed by the loss by an order of magnitude from the baseline.
For forward (Loss) and backward (Gradient) passes combined, \ours is within \qty{3}{MB} of the lowest possible memory consumption.
Compared to Gemma 2 (\qty{2}{B}) all these models have a smaller ratio of the vocabulary size to hidden dimension. This has two impacts. 

First, the number of tokens that have a significant gradient is largely constant (it is dependent on the data type). Therefore proportionally less of the gradient will be filtered out. 

Second, for all other methods increasing the hidden dimension increase the amount of parallelism that can be achieved.  Liger Kernels~\citep{liger2024} sets its chunk size based on $\nicefrac{|V|}{D}$ -- the lower that ratio, the bigger the chunk size. As $\nicefrac{|V|}{D}$ continues to decrease, Liger Kernels is able to make better use of the GPU. All other methods use two matrix multiplications to compute the gradient. The amount of work that can be performed in parallel to compute $\nabla \mE$ and $\nabla \mC$ is $B \times D$ and $|V| \times D$, respectively\footnote{Ignoring split-k matrix multiplication kernels for simplicity.}. The amount of parallel work for \ours is $B \times |V|$, thus increasing $D$ increases the amount of work but not the amount of parallelism. It may be possible leverage ideas from split-k matrix multiplication kernels to expose more parallelism to \ours for large values of $D$.

For the smallest $\nicefrac{|V|}{D}$ considered, Phi 3.5 Mini ($|V|$=\num{32064}, D=\num{3072}) ours is approximately $50\%$ slower (\qty{12}{ms}) than \texttt{torch.compile} (although it uses substantially less memory). In our experiments, this increase in linear-cross-entropy loss computation time is largely negligible and only increases training time by one to two percent.

\begin{figure}
    \captionsetup[subfigure]{justification=centering}
    \begin{subfigure}{0.95\linewidth}
    \centering
    \includegraphics[width=\linewidth]{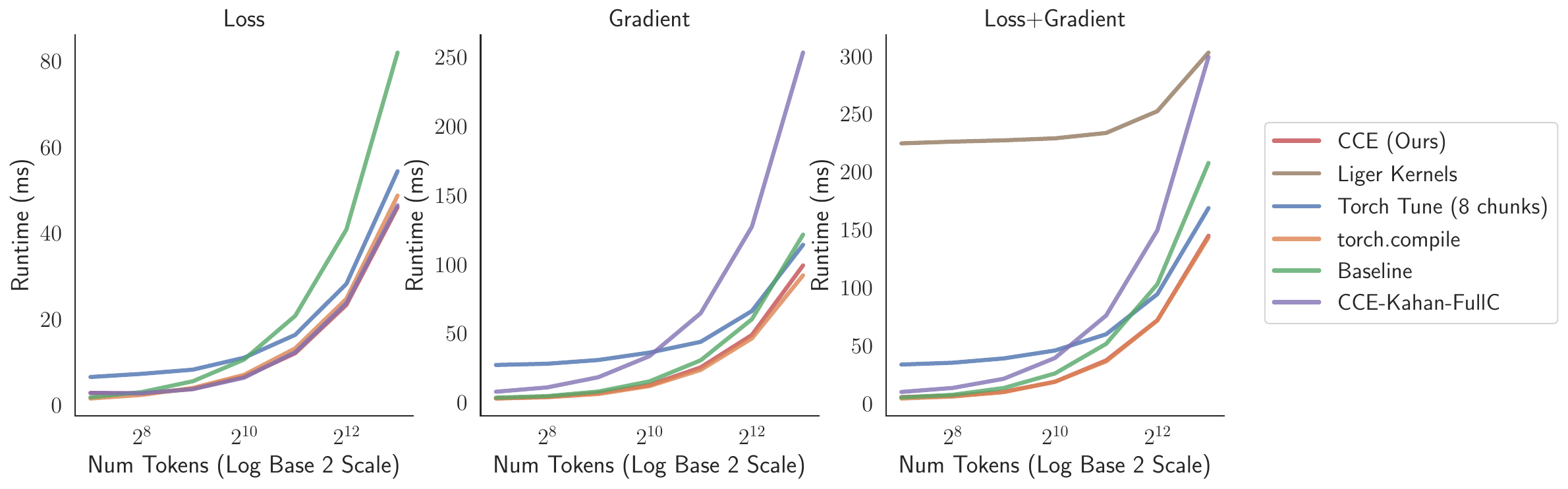}
    \caption{Gemma 2 \qty{2}{B}}
    \end{subfigure}
    \begin{subfigure}{0.95\linewidth}
    \centering
    \includegraphics[width=\linewidth]{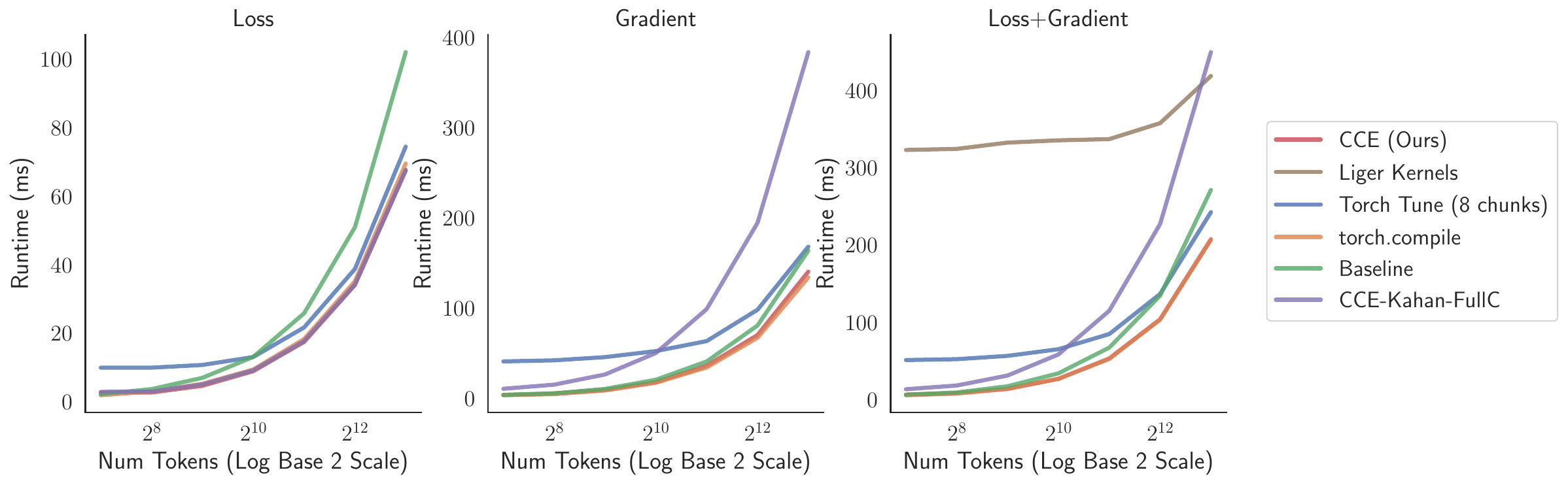}
    \caption{Gemma 2 \qty{9}{B}}
    \end{subfigure}
    \begin{subfigure}{0.95\linewidth}
    \centering
    \includegraphics[width=\linewidth]{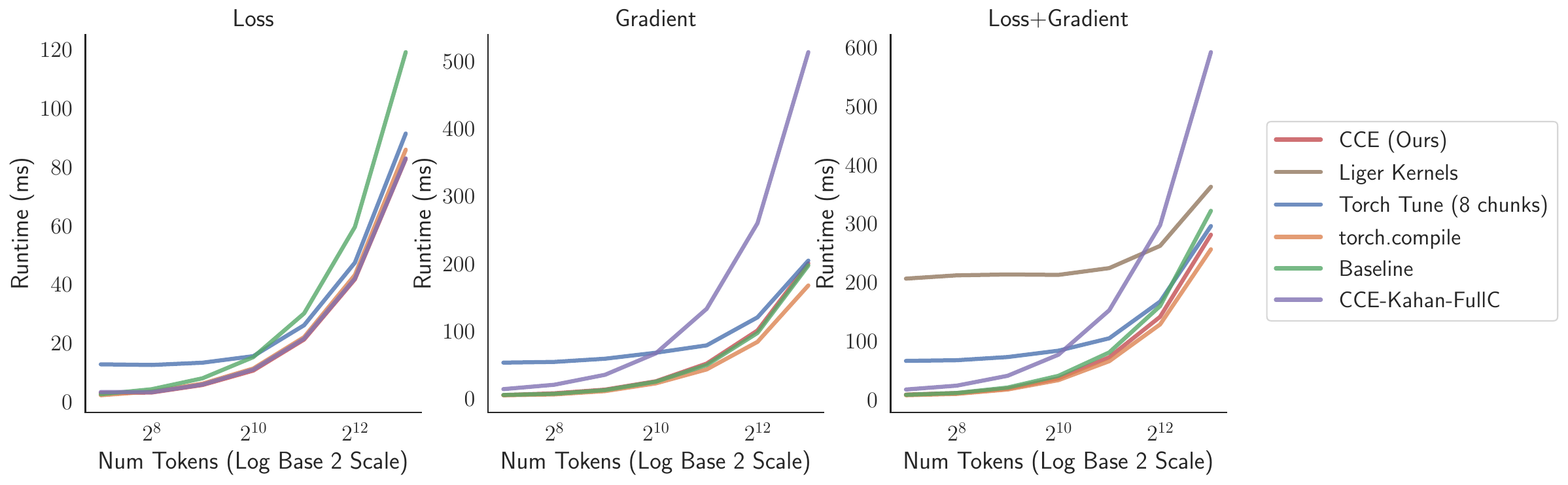}
    \caption{Gemma 2 \qty{27}{B}}
    \end{subfigure}
    \begin{subfigure}{0.95\linewidth}
    \centering
    \includegraphics[width=\linewidth]{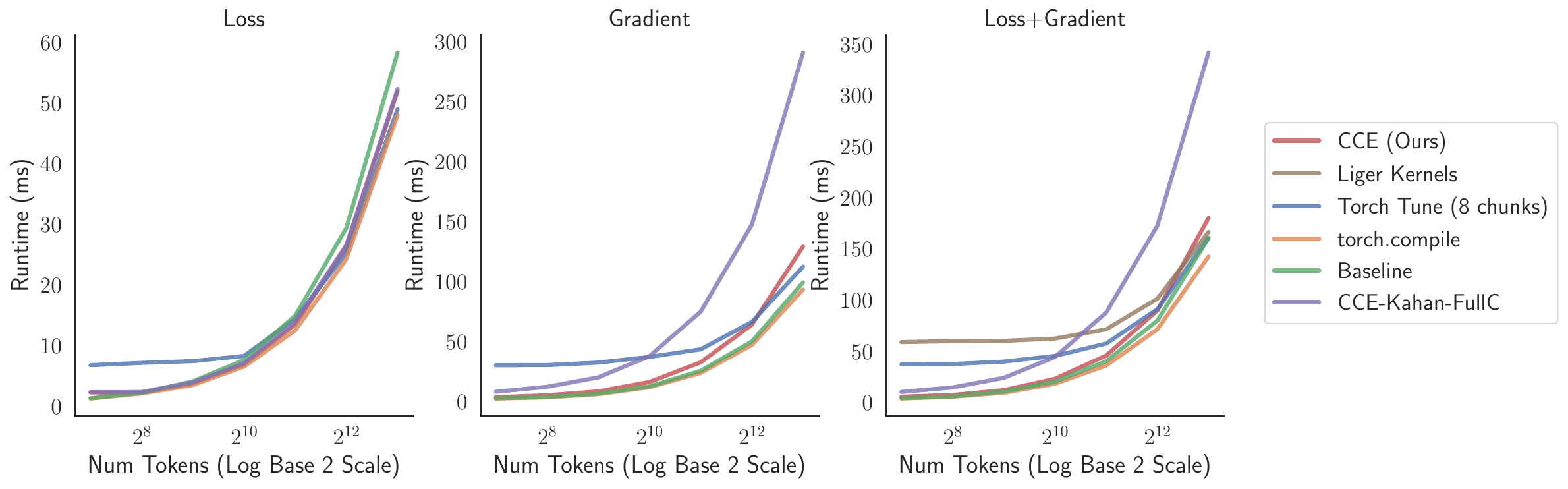}
    \caption{Mistral NeMo}
    \end{subfigure}
     \caption{Performance of \ours and baselines for all models with a varying batch sizes. Results averaged over 5 seeds. Continued in \cref{fig:perf-vs-tokens-part-2}.}
     \label{fig:perf-vs-tokens-part-1}
\end{figure}

\begin{figure}
    \captionsetup[subfigure]{justification=centering}
    \begin{subfigure}{0.95\linewidth}
    \centering
    \includegraphics[width=\linewidth]{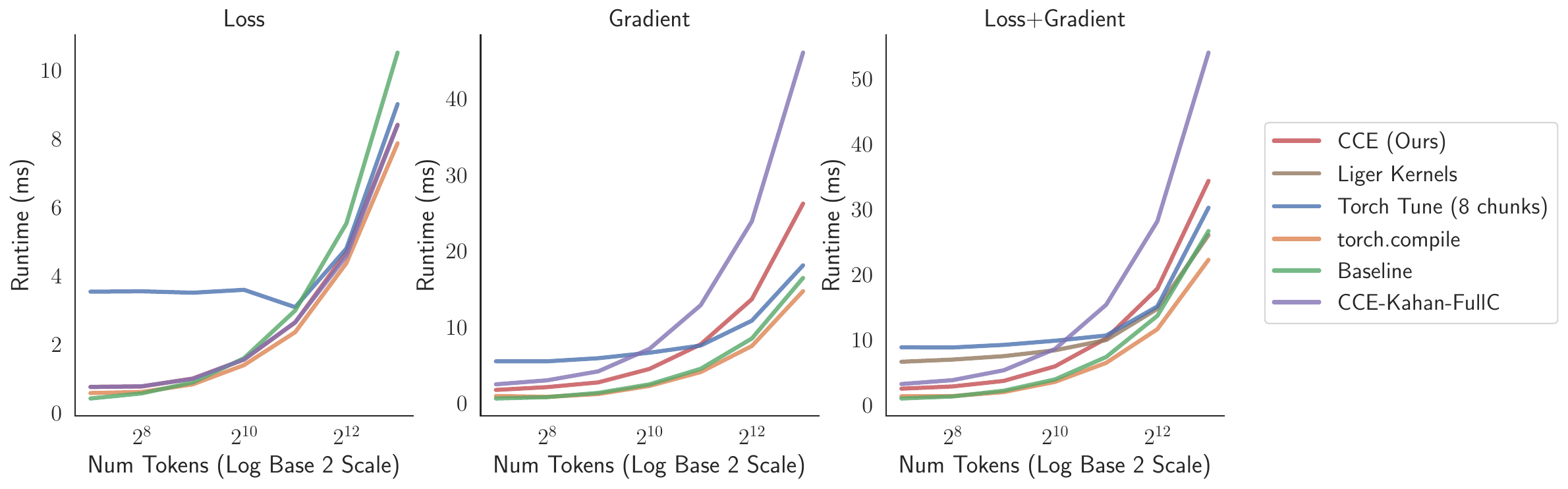}
    \caption{Phi 3.5 Mini}
    \end{subfigure}
    \begin{subfigure}{0.95\linewidth}
    \centering
    \includegraphics[width=\linewidth]{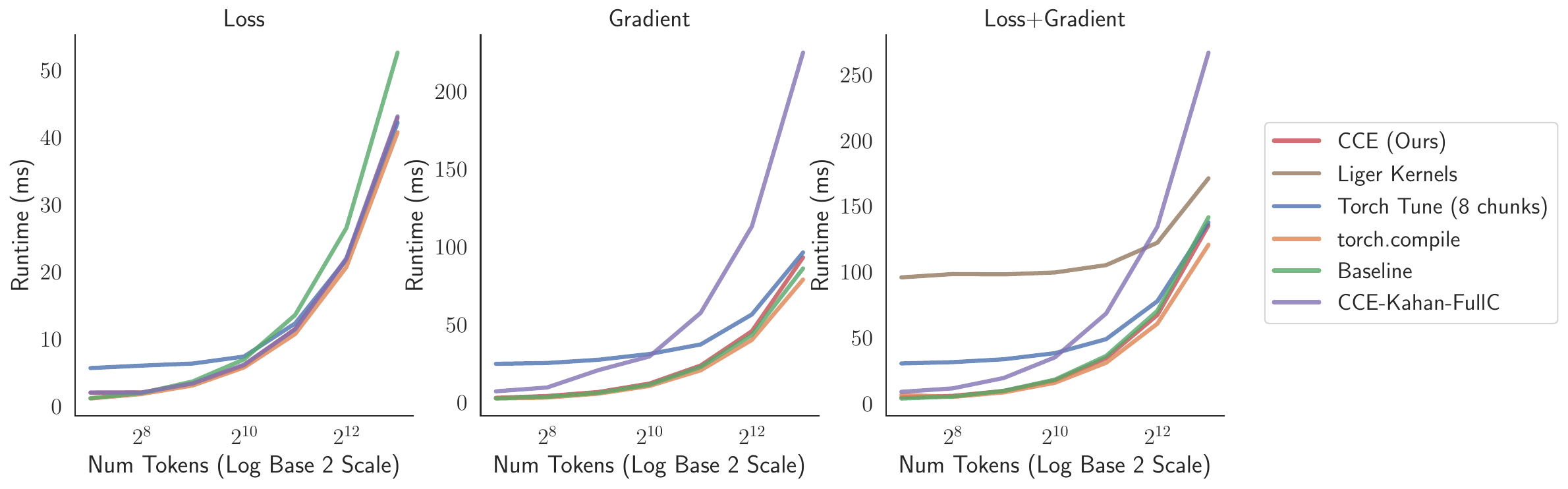}
    \caption{Qwen 2.5 \qty{7}{B}}
    \end{subfigure}
    \begin{subfigure}{0.95\linewidth}
    \centering
    \includegraphics[width=\linewidth]{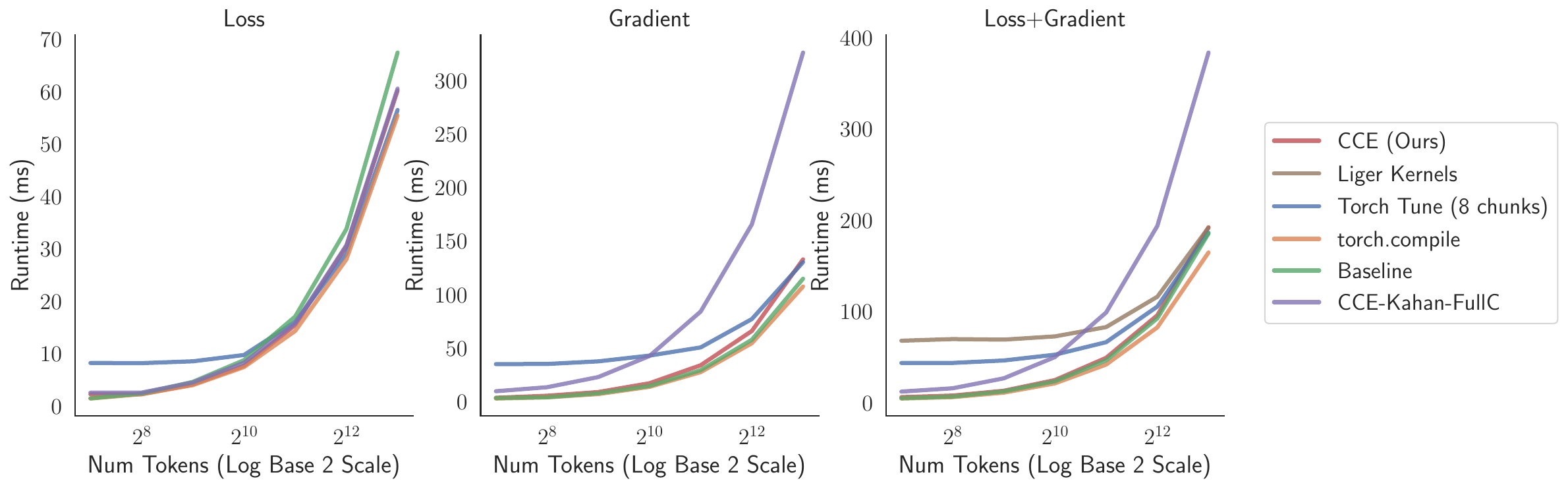}
    \caption{Qwen 2.5 \qty{32}{B}}
    \end{subfigure}
    \caption{Performance of \ours and baselines for all models with a varying batch sizes. Results averaged over 5 seeds.}
    \label{fig:perf-vs-tokens-part-2}
\end{figure}

We also consider how changing the number of tokens changes performance (\cref{fig:perf-vs-tokens-part-1,fig:perf-vs-tokens-part-2}). We find that \ours behaves very similarly to Baseline and \texttt{torch.compile}. Further, because \ours does not utilize chunking, it does not reach a point where the overhead of dispatching all the kernels becomes the dominating factor. We also find that while \oursfullc is slower than the Liger Kernel and Torch Tune baselines with a large number of tokens, it becomes more performant than those baselines as the number of tokens reduces.

\section{Memory use method details}
\cref{tab:teaser-data} contains the raw numbers used to create \cref{fig:teaser}. The maximum batch size for 16 GPUs was calculated by assuming that the total amount of memory available is $75 \times 16$ (\ie each \qty{80}{GB} GPU will be fully occupied expect for a \qty{5}{GB} buffer for various libraries), then subtracting the memory used for weights + optimizer + gradients and then diving by the memory used per token.

The numbers in \cref{tab:teaser-data} are computed using the following methods. When present, the number of tokens is assumed to be \num{65536}.

We compute the amount of memory used for intermediate activations as the number of layers times the hidden size times number of tokens times 2 bytes per bfloat16. This assumes the use of activation/gradient checkpointing~\citep{chen2016checkpointing} for transformer layer. 

The amount of memory used by the logits is the number of tokens times the vocabulary size times 4 bytes per float32. This likely undercounts the amount of memory used for computing the probability distribution, as its common to also keep a copy of the logits in bfloat16 and, for models like Gemma 2~\citep{gemma2} that use logit softcapping, an additional copy of the logits after softcapping may be needed. However, this method can be uniformly applied to all models.

The amount of memory used by Weights+Opt+Grad is the number of parameters times 4 (parameters, gradient, and Adam first and second moments) times 2 bytes per bfloat16.

\section{Floating Point Addition}

Here we provide a brief explanation of floating point addition and how it relates to our proposed gradient filtering.

Given two numbers $a$ and $b$ represented using floating point, such that $|a| < |b|$, the following steps are performed

\begin{enumerate}
    \item Separate the mantissa (the fractional part) and the exponent from both numbers $a$ and $b$.
    \item Re-write the mantissa of the smaller number ($a$ in our case) such that it shares the same exponent as the $b$.
    \item Add the re-written mantissa of $a$ to the mantissa of $b$.
    \item Combine the resulting mantissa and exponent of $b$ and then convert them into normalized form.
\end{enumerate}

Step 2 is where truncation happens and the intuition of gradient filtering comes from. In bfloat16, if the exponent of $b$ is more than $2^7$ times larger than that of a, the 7-bit mantissa no longer has enough precision to represent any of $a$'s mantissa and in the process of re-writing, $a$ will be, in effect, set to zero. For gradient filtering, we are only concerned with values in the range $[0, 1]$, so the threshold of $2^{-12}$ means that we only keep values that don’t get rounded to zero when b = $2^{-5}$.

\begin{table}
    \centering
    \setlength{\tabcolsep}{2pt}
    \resizebox{0.975\textwidth}{!}{
    \begin{tabular}{lc rc rc rc rc rc r}
    \toprule
    && \multicolumn{3}{c}{Loss} &&  \multicolumn{3}{c}{Gradient} && \multicolumn{3}{c}{Loss+Gradient} \\
    \cmidrule{3-5}
    \cmidrule{7-9} \cmidrule{11-13}
    Method && Memory && Time  && Memory  && Time && Memory && Time  \\
    \midrule
    Gemma 2 (\qty{9}{B})~\citep{gemma2} ($|V|$=\num{256000}, D=\num{3584}) \\
    Lower bound && \qty{0.004}{MB} && && \qty{1806}{MB} &&  && \qty{1806}{MB} \\[0.5em]
    \ours (Ours) && \textbf{\qty{1}{MB}} && \textbf{\qty{68}{ms}} && \textbf{\qty{1808}{MB}} && \qty{141}{ms} && \textbf{\qty{1809}{MB}} && \qty{208}{ms} \\
    Liger Kernels~\citep{liger2024} && \qty{2119}{MB} && \qty{418}{ms} &&  && && \qty{2119}{MB} && \qty{419}{ms} \\
    \citet{torchtune2024} (8 chunks) && \qty{8000}{MB} && \qty{75}{ms} && \qty{3264}{MB} && \qty{168}{ms} && \qty{11264}{MB} && \qty{243}{ms} \\
    \texttt{torch.compile} && \qty{4000}{MB} && \qty{70}{ms} && \qty{12000}{MB} && \textbf{\qty{134}{ms}} && \qty{16000}{MB} && \textbf{\qty{207}{ms}} \\
    Baseline && \qty{24000}{MB} && \qty{102}{ms} && \qty{16000}{MB} && \qty{164}{ms} && \qty{28000}{MB} && \qty{271}{ms} \\[0.5em]
    \oursfullc && \qty{1}{MB} && \qty{68}{ms} && \qty{3558}{MB} && \qty{384}{ms} && \qty{3559}{MB} && \qty{450}{ms} \\
    \midrule
    Gemma 2 (\qty{27}{B})~\citep{gemma2} ($|V|$=\num{256000}, D=\num{4608}) \\
    Lower bound && \qty{0.004}{MB} && && \qty{2322}{MB} &&  && \qty{2322}{MB} \\[0.5em]
    \ours (Ours) && \textbf{\qty{1}{MB}} && \textbf{\qty{83}{ms}} && \textbf{\qty{2324}{MB}} && \qty{200}{ms} && \textbf{\qty{2325}{MB}} && \qty{281}{ms} \\
    Liger Kernels~\citep{liger2024} && \qty{2948}{MB} && \qty{361}{ms} &&  && && \qty{2948}{MB} && \qty{363}{ms} \\
    \citet{torchtune2024} (8 chunks) && \qty{8000}{MB} && \qty{91}{ms} && \qty{4768}{MB} && \qty{204}{ms} && \qty{12768}{MB} && \qty{296}{ms} \\
    \texttt{torch.compile} && \qty{4000}{MB} && \qty{86}{ms} && \qty{12000}{MB} && \textbf{\qty{168}{ms}} && \qty{16000}{MB} && \textbf{\qty{256}{ms}} \\
    Baseline && \qty{24000}{MB} && \qty{119}{ms} && \qty{16000}{MB} && \qty{197}{ms} && \qty{28000}{MB} && \qty{322}{ms} \\[0.5em]
    \oursfullc && \qty{1}{MB} && \qty{83}{ms} && \qty{4574}{MB} && \qty{513}{ms} && \qty{4575}{MB} && \qty{593}{ms} \\
    \midrule
    Mistral NeMo~\citep{mistralnemo} ($|V|$=\num{131072}, D=\num{5120}) \\
    Lower bound && \qty{0.004}{MB} && && \qty{1360}{MB} &&  && \qty{1360}{MB} \\[0.5em]
    \ours (Ours) && \textbf{\qty{0.6}{MB}} && \qty{52}{ms} && \textbf{\qty{1361}{MB}} && \qty{129}{ms} && \textbf{\qty{1362}{MB}} && \qty{180}{ms} \\
    Liger Kernels~\citep{liger2024} && \qty{1872}{MB} && \qty{166}{ms} &&  && && \qty{1872}{MB} && \qty{167}{ms} \\
    \citet{torchtune2024} (8 chunks) && \qty{2048}{MB} && \qty{49}{ms} && \qty{3348}{MB} && \qty{113}{ms} && \qty{5396}{MB} && \qty{161}{ms} \\
    \texttt{torch.compile} && \qty{2048}{MB} && \textbf{\qty{48}{ms}} && \qty{6144}{MB} && \textbf{\qty{94}{ms}} && \qty{8192}{MB} && \textbf{\qty{143}{ms}} \\
    Baseline && \qty{10240}{MB} && \qty{58}{ms} && \qty{8192}{MB} && \qty{100}{ms} && \qty{12288}{MB} && \qty{161}{ms} \\[0.5em]
    \oursfullc && \qty{0.6}{MB} && \qty{52}{ms} && \qty{2641}{MB} && \qty{291}{ms} && \qty{2642}{MB} && \qty{342}{ms} \\
    \midrule
    Phi 3.5 Mini~\citep{phi35} ($|V|$=\num{32064}, D=\num{3072}) \\
    Lower bound && \qty{0.004}{MB} && && \qty{236}{MB} &&  && \qty{236}{MB} \\[0.5em]
    \ours (Ours) && \textbf{\qty{0.2}{MB}} && \qty{8}{ms} && \textbf{\qty{236}{MB}} && \qty{26}{ms} && \textbf{\qty{236}{MB}} && \qty{34}{ms} \\
    Liger Kernels~\citep{liger2024} && \qty{487}{MB} && \qty{26}{ms} &&  && && \qty{488}{MB} && \qty{26}{ms} \\
    \citet{torchtune2024} (8 chunks) && \qty{502}{MB} && \qty{9}{ms} && \qty{451}{MB} && \qty{18}{ms} && \qty{953}{MB} && \qty{30}{ms} \\
    \texttt{torch.compile} && \qty{502}{MB} && \textbf{\qty{8}{ms}} && \qty{1504}{MB} && \textbf{\qty{15}{ms}} && \qty{2006}{MB} && \textbf{\qty{22}{ms}} \\
    Baseline && \qty{2506}{MB} && \qty{11}{ms} && \qty{2004}{MB} && \qty{16}{ms} && \qty{3006}{MB} && \qty{27}{ms} \\[0.5em]
    \oursfullc && \qty{0.2}{MB} && \qty{8}{ms} && \qty{424}{MB} && \qty{46}{ms} && \qty{424}{MB} && \qty{54}{ms} \\
    \midrule
    Qwen 2.5 (\qty{7}{B})~\citep{qwen2.5} ($|V|$=\num{152064}, D=\num{3584}) \\
    Lower bound && \qty{0.004}{MB} && && \qty{1096}{MB} &&  && \qty{1096}{MB} \\[0.5em]
    \ours (Ours) && \textbf{\qty{0.6}{MB}} && \qty{43}{ms} && \textbf{\qty{1098}{MB}} && \qty{93}{ms} && \textbf{\qty{1097}{MB}} && \qty{136}{ms} \\
    Liger Kernels~\citep{liger2024} && \qty{1394}{MB} && \qty{171}{ms} &&  && && \qty{1394}{MB} && \qty{171}{ms} \\
    \citet{torchtune2024} (8 chunks) && \qty{2379}{MB} && \qty{42}{ms} && \qty{2540}{MB} && \qty{96}{ms} && \qty{4921}{MB} && \qty{138}{ms} \\
    \texttt{torch.compile} && \qty{2376}{MB} && \textbf{\qty{41}{ms}} && \qty{7128}{MB} && \textbf{\qty{79}{ms}} && \qty{9504}{MB} && \textbf{\qty{121}{ms}} \\
    Baseline && \qty{11880}{MB} && \qty{53}{ms} && \qty{9504}{MB} && \qty{86}{ms} && \qty{14256}{MB} && \qty{142}{ms} \\[0.5em]
    \oursfullc && \qty{0.6}{MB} && \qty{43}{ms} && \qty{2138}{MB} && \qty{225}{ms} && \qty{2138}{MB} && \qty{267}{ms} \\
    \midrule
    Qwen 2.5 (\qty{32}{B})~\citep{qwen2.5} ($|V|$=\num{152064}, D=\num{5120}) \\
    Lower bound && \qty{0.004}{MB} && && \qty{1565}{MB} &&  && \qty{1565}{MB} \\[0.5em]
    \ours (Ours) && \textbf{\qty{0.6}{MB}} && \qty{60}{ms} && \textbf{\qty{1566}{MB}} && \qty{133}{ms} && \textbf{\qty{1567}{MB}} && \qty{193}{ms} \\
    Liger Kernels~\citep{liger2024} && \qty{2159}{MB} && \qty{192}{ms} &&  && && \qty{2161}{MB} && \qty{192}{ms} \\
    \citet{torchtune2024} (8 chunks) && \qty{2376}{MB} && \qty{57}{ms} && \qty{3882}{MB} && \qty{130}{ms} && \qty{6259}{MB} && \qty{186}{ms} \\
    \texttt{torch.compile} && \qty{2376}{MB} && \textbf{\qty{56}{ms}} && \qty{7128}{MB} && \textbf{\qty{108}{ms}} && \qty{9504}{MB} && \textbf{\qty{165}{ms}} \\
    Baseline && \qty{11880}{MB} && \qty{68}{ms} && \qty{9504}{MB} && \qty{115}{ms} && \qty{14256}{MB} && \qty{186}{ms} \\[0.5em]
    \oursfullc && \qty{0.6}{MB} && \qty{61}{ms} && \qty{3052}{MB} && \qty{326}{ms} && \qty{3053}{MB} && \qty{384}{ms} \\
    \bottomrule
    \end{tabular}}
    \caption{Memory usage and time of \ours, Liger Kernels, Torch Tune, \texttt{torch.compile}, and Baseline for additional models. Batch of \num{8192} tokens. Results averaged over 5 seeds.}
    \label{tab:additional-results}
\end{table}


\begin{table}
    \centering
    \setlength{\tabcolsep}{4pt}
    \resizebox{0.975\textwidth}{!}{
    \begin{tabular}{l r r r r r r}
    \toprule
    Model & Logits & Activations & Weights+Opt+Grad & Max Batch Size (Before) & Max Batch Size (After) & Increase \\
    \midrule
    GPT 2 & \qty{12564}{MB} & \qty{1152}{MB} & \qty{1045}{MB} & \num{5866190} & \num{69845595} & \num{11.9}$\times$ \\
    GPT Neo (\qty{1.3}{B}) & \qty{12564}{MB} & \qty{6144}{MB} & \qty{10421}{MB} & \num{4268047} & \num{12996042} & \num{3.0}$\times$ \\
    GPT Neo (\qty{2.7}{B}) & \qty{12564}{MB} & \qty{10240}{MB} & \qty{20740}{MB} & \num{3471784} & \num{7731585} & \num{2.2}$\times$ \\
    Gemma (\qty{2}{B}) & \qty{64000}{MB} & \qty{4608}{MB} & \qty{19121}{MB} & \num{1155515} & \num{17204330} & \num{14.9}$\times$ \\
    Gemma 2 (\qty{27}{B}) & \qty{64000}{MB} & \qty{26496}{MB} & \qty{207727}{MB} & \num{739448} & \num{2525554} & \num{3.4}$\times$ \\
    Gemma 2 (\qty{2}{B}) & \qty{64000}{MB} & \qty{7488}{MB} & \qty{19946}{MB} & \num{1108206} & \num{10580057} & \num{9.5}$\times$ \\
    Llama 2 (\qty{13}{B}) & \qty{8000}{MB} & \qty{25600}{MB} & \qty{99303}{MB} & \num{2203057} & \num{2891512} & \num{1.3}$\times$ \\
    Llama 2 (\qty{7}{B}) & \qty{8000}{MB} & \qty{16384}{MB} & \qty{51410}{MB} & \num{3164429} & \num{4709560} & \num{1.5}$\times$ \\
    Llama 3 (\qty{70}{B}) & \qty{32064}{MB} & \qty{81920}{MB} & \qty{538282}{MB} & \num{397019} & \num{552414} & \num{1.4}$\times$ \\
    Llama 3 (\qty{8}{B}) & \qty{32064}{MB} & \qty{16384}{MB} & \qty{61266}{MB} & \num{1579333} & \num{4670136} & \num{3.0}$\times$ \\
    Mistral \qty{7}{B} & \qty{8000}{MB} & \qty{16384}{MB} & \qty{55250}{MB} & \num{3154108} & \num{4694200} & \num{1.5}$\times$ \\
    Mixtral 8x\qty{7}{B} & \qty{8000}{MB} & \qty{16384}{MB} & \qty{356314}{MB} & \num{2344949} & \num{3489944} & \num{1.5}$\times$ \\
    Phi 1.5 & \qty{12574}{MB} & \qty{6144}{MB} & \qty{10821}{MB} & \num{4264482} & \num{12991781} & \num{3.0}$\times$ \\
    Phi 3 Medium & \qty{8003}{MB} & \qty{25600}{MB} & \qty{106508}{MB} & \num{2188824} & \num{2873067} & \num{1.3}$\times$ \\
    Qwen 1.5 (\qty{7}{B}) & \qty{37912}{MB} & \qty{16384}{MB} & \qty{58909}{MB} & \num{1412087} & \num{4679564} & \num{3.3}$\times$ \\
    \bottomrule
    \end{tabular}}
    \caption{Raw data for \cref{fig:teaser}.  Memory usage calculated using a global batch size of \num{65536}.}
    \label{tab:teaser-data}
\end{table}

\begin{algorithm}[t]
    \begin{tabularx}{\textwidth}{lX}
      \textbf{Inputs:} & $\mat{\mE} \in \mathbb{R}^\dim{D}{N}$, $\mat{\mC} \in \mathbb{R}^\dim{D}{|V|}$, $\vec{\mLSE} \in \mathbb{R}^N$, $\nabla \vec{\mCEL} \in \mathbb{R}^N$, and $\vec{x} \in \mathbb{R}^N$.\\
      & Block sizes $\BN$, $\BV$, and $\BD$.\\
      & Accuracy threshold $\varepsilon$.\\
      & $\vec{v} = [1, \ldots, \V]$.\\
      \textbf{Outputs:} & $\nabla\mat{\mE} \in \mathbb{R}^\dim{D}{N}$, $\nabla\mat{\mC} \in \mathbb{R}^\dim{D}{|V|}$
    \end{tabularx}
    \vspace{0.25em}
    \hrule
    \vspace{0.25em}
    \begin{algorithmic}
\For{all pairs of blocks $\mat{\mE}_n$, $\mat{\mC}_v$}\Comment{Divide $\mat{\mE}$ and $\mat{\mC}$ into blocks of size $\dim{D}{\BN}$ and $\dim{D}{\BV}$}
        \State $\mat{A}_{nv} = \vec{0}_\dim{\BV}{\BN}$\Comment{Zero matrix of size $\dim{\BV}{\BN}$ in on-chip SRAM}
        \For{blocks $\mat{\mE}_{n,d}$, $\mat{\mC}_{v,d}$}\Comment{Divide $\mat{\mE}_n$ and $\mat{\mC}_v$ into blocks of $\dim{\BD}{\BN}$ and $\dim{\BD}{\BV}$}
            \State $\mat{A}_{nv} \pluseq \mat{\mC}_{v,d}^\top \cdot \mat{\mE}_{n,d}$ \Comment{Blockwise matrix multiplication}
        \EndFor
        \State $\mat{S}_{nv} = \exp(\mat{A}_{nv} - \vec{\mLSE}_n)$ \Comment{Compute the softmax}
        \State $\mat{G}_{nv} = \left[\left[\vec{v}_v = \vec{x}^\top_n\right]\right] - \mat{S}_{nv} $ \Comment{Gradient of cross-entropy loss wrt. logits}
        \If{all$(|\mat{G}_{nv}| < \varepsilon)$}
            \State \textbf{skip} \Comment{Skip computation if below desired numerical precision}
        \EndIf
        \For{blocks $\mat{\mE}_{n,d}$, $\mat{\mC}_{v,d}$}\Comment{Divide $\mat{\mE}_n$ and $\mat{\mC}_m$ into blocks of $\dim{\BD}{\BN}$ and $\dim{\BD}{\BV}$}
            \State $\nabla \mat{\mE}^\top_{n,d} \pluseq \left(\mat{G}_{nv} \cdot \nabla \vec{\mCEL}_n\right) \mat{\mC}_{v,d}$\Comment{Locking thread-safe gradient update}
            \State $\nabla \mat{\mC}^\top_{v,d} \pluseq \left(\mat{G}_{nv} \cdot \nabla \vec{\mCEL}_n\right)^\top \mat{\mE}_{n,d}$\Comment{Locking thread-safe gradient update}
        \EndFor
    \EndFor
    \end{algorithmic}
    \caption{Memory-efficient linear-cross-entropy loss, backward pass}
    \lblalg{cel_bck}
\end{algorithm}

\end{document}